\documentclass[a4paper,11pt,final]{article}

\usepackage{graphicx,amsmath,amssymb,algorithm,algorithmic} 
\usepackage{epsfig}

\begin{document}

\title{ \LARGE\bf Designing color symmetry  in stigmergic art}

\author{Hendrik Richter \\
HTWK Leipzig University of Applied Sciences \\ Department of
Electrical Engineering and Information Technology\\
        Postfach 301166, D--04251 Leipzig, Germany. \\ Email: 
hendrik.richter@htwk-leipzig.de. }

\maketitle
\begin{abstract} Color symmetry is an extension of symmetry imposed by isometric transformations and means that the colors of geometrical objects are assigned according to the symmetry properties of the objects. A color symmetry permutes the coloring of the objects consistently with their symmetry group. We apply this concept to bio-inspired generative art. Therefore, we interpret the geometrical objects as motifs that may repeat themselves with a symmetry-consistent coloring. The motifs are obtained by design principles from stigmergy. We discuss a design procedure and present visual results.
\end{abstract}

\section{Introduction}
Symmetry is a major organizational principle 
in different kinds of design~\cite{conway08,weil52}. The principle is present in living matter formed by evolutionary processes which frequently possesses a high degree of symmetry. It  equally applies to man-made objects, whether it be in engineering, where symmetry is economical due to the use of identical components or parts, and also in art, where playing with symmetry is a way of expressing aesthetic concepts.

Bio-inspired generative art is a field in evolutionary computation dealing with employing algorithmic ideas and templates observed in biology for creating artworks~\cite{green15,jac07,neu20}. As symmetry is both an important aesthetic concept in art and an organizational principle in living matter, it follows rather naturally  that  different forms and meanings of symmetry matter to bio-inspired generative art, see e.g.~\cite{al17,green12,rich20} for explicitly addressing the topic. 
In this paper, we focus on a special kind of symmetry: color symmetry. 
Assume that  geometrical objects have symmetry properties, which can be described by their symmetry group. If the objects can be colored, then color symmetry implies that the colors of the objects are derived consistently from the symmetry group~\cite{cox86,rich20,roth82,schwarz84,sen83,sen88}.    
 A particularly straightforward way of obtaining symmetry properties frequently employed in visual art is to interpret the geometrical objects as motifs which may repeat themselves. In this case 
color symmetry means a symmetry-consistent coloring of each motif and its copies.

 \begin{figure}[tb]
 
\center
\includegraphics[trim = 80mm 40mm 70mm 0mm,clip,width=8.35cm, height=7.5cm]{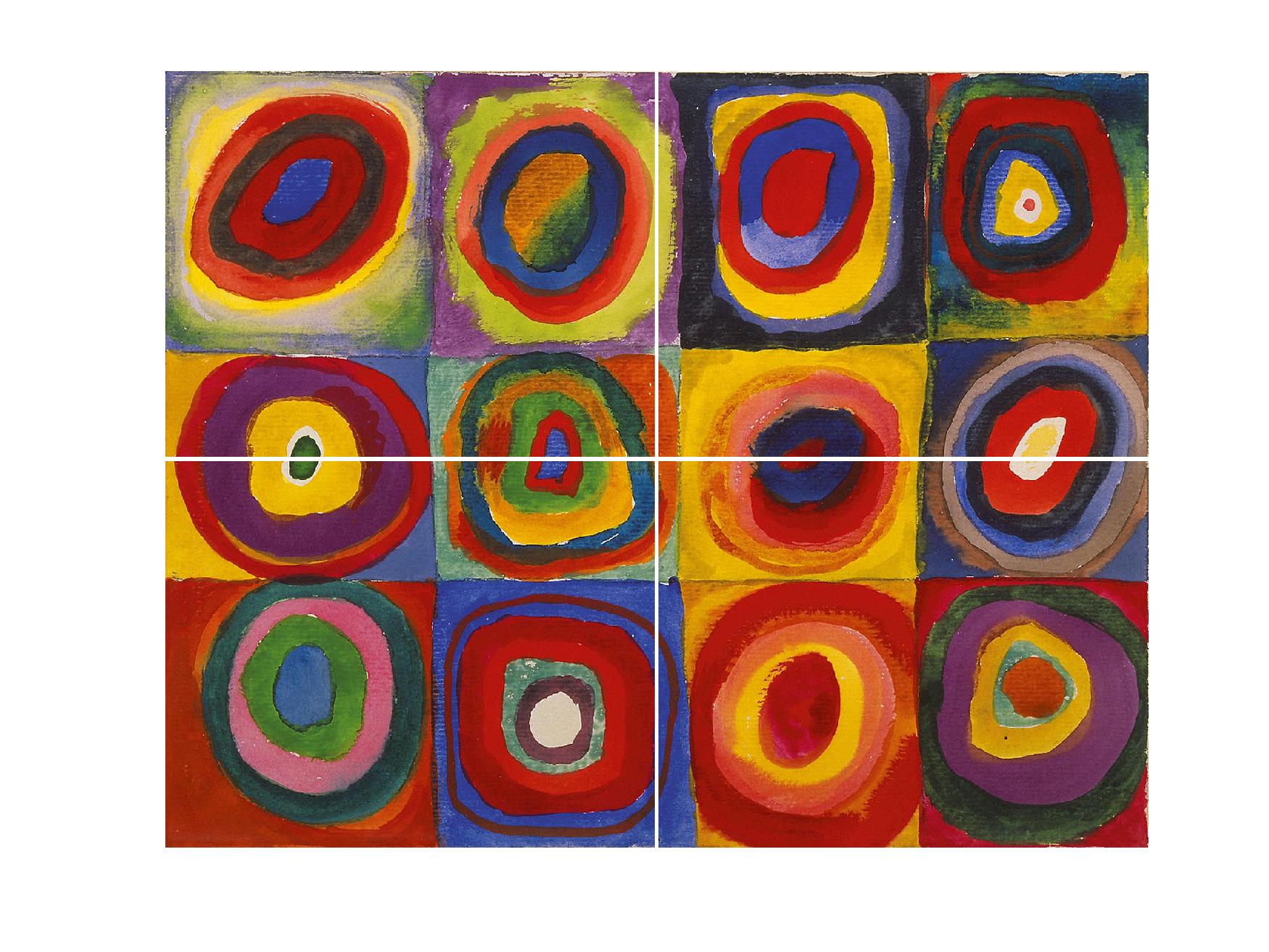}
\includegraphics[trim = 80mm 40mm 70mm 0mm,clip,width=8.35cm, height=7.5cm]{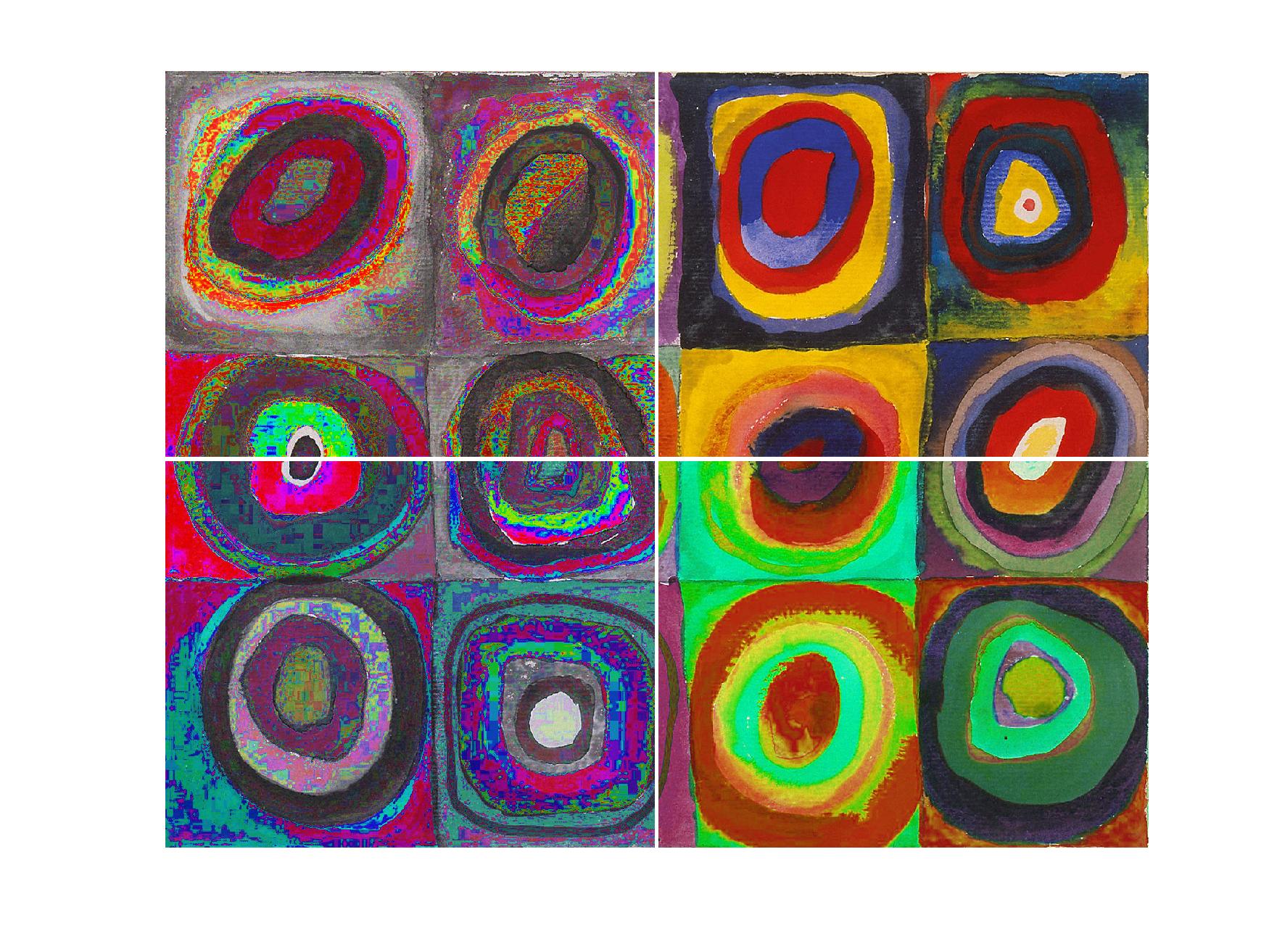}

\hspace{1.8cm}(a) \hspace{3.5cm} (b)

\caption{\textbf{(a)} Vasily Kandinsky (1866-1944): {\it Farbstudie - Quadrate mit konzentrischen Ringen} (Color study. Squares with concentric rings),
1913. \textbf{(b)} Color symmetry using a color permutation associated with the symmetry group of the rectangle.
}
\label{fig:kandins}
\end{figure}

An illustrative example from the history of art for exploring the relationships between geometrical objects, motifs and colors is 
  Vasily Kandinsky's {\it Farbstudie - Quadrate mit konzentrischen Ringen} (Color study. Squares with concentric rings), (1913)~\cite{beck07}.  
 Fig. \ref{fig:kandins}a shows the study, which consists of 12 squares 
filling a rectangle in 3 rows with each square containing a central circle surrounded by concentric
rings. Some of the central circles are slightly off-center or almost squarish and all are surrounded by
concentric circles of different colors.  The colors are mainly the primary and the secondary
colors of the RYB (red-yellow-blue) color wheel~\cite{itten73,rhyne17}, but there are also various tertiary colors.  Sometimes the circles' colors repeat themselves, some other circles traverse a color space. For instance, we have the sequence (from the inside out) red-blue-red-yellow-blue-black in the upper square second to the right.  The arrangement can be interpreted as 12 iterations (but also as more or less faithful copies) of a motif. For a strict symmetry 
we might expect the geometrical form and/or the color of a constant element within the motif to differ each time, or to stay constant all the time, but this was not Kandinsky's intention. 
 It is rather an exploration of motifs taken from basic geometrical forms (squares and circles). In addition, we have variations in the form and color within individual
copies and throughout the study experiments with different hue and brightness. Nevertheless, we may see Kandinsky's color study  as anticipating  and preceding some elements of the conceptual framework of color symmetry.

Similarly to previous works which used Kandinsky's art as an
 inspiration of generative art~\cite{neu17,neu20,zhang16}, the design of color symmetry discussed in this paper also gives a reinterpretation of the color study shown in Fig. \ref{fig:kandins}a. The
motif generation  is done using  algorithms implementing a stigmergic design.  
Stigmergy is a principle of self-organization in nature, particularly in swarms, see e.g.~\cite{ al13,green12,green14,mou07,urb05,urb11} for applications in bio-inspired generative art.

The paper is structured as follows. After these prefatory remarks,  Sec. \ref{sec:color} 
 briefly reviews meanings and concepts of color, symmetry and color symmetry in visual art. We also discuss how color symmetry can be imposed by color permutations.
Sec. \ref{sec:stig} recalls stigmergic generation of geometrical objects and particularly shows the motif design used in the paper. Numerical experiments and visual results are given in Sec. \ref{sec:exp}, before concluding remarks close the paper with a summary and pointers to future work.

\section{Color, symmetry, and color symmetry} \label{sec:color}
For describing the significance and meaning of color in art, there is the dictum of the Hungarian-French 20th century painter and graphic designer Victor Vasarely 
 that 
``every form is a base for color, every color is the attribute of a form.'' Put differently, if we see
visual art as composed of geometrical objects, then color can be viewed as basically a property of these objects. Moreover, if the artistic content of the artwork relates to how and where the geometrical objects are arranged in the image, then different colors allow to distinguish and contrast the objects.   Suppose there is a point $P_i=(x_i,y_i)$ from a point set
in a two-dimensional $(x,y)$-plane. Then, a color $c_i$ is an attribute (or property) of the point $P_i$. In other words, a colored point is specified by $P_i=(x_i,y_i,c_i)$. Geometrical objects are built from a set of points. The objects are (monochromatically) colored for all points of the set having the same color.

A geometrical object may possess symmetry, which implies that the object is invariant under certain transformations~\cite{conway08,liu10,martin82,weil52}. The set of transformations under which the geometrical object is symmetric yields the symmetry group $\mathcal{G}$ of the object. On the one hand, we may focus on the symmetry properties of a single geometrical object. For creating or analyzing images, on the other hand, we may consider a geometrical object (which may or may not have symmetry properties itself)  as a motif. In an image such a motif may appear in several copies, and we can study the symmetry properties between these copies.

\begin{figure}[tb]
 
\centering
\includegraphics[trim = 220mm 160mm 180mm 140mm,clip,width=6.8cm, height=7.2cm]{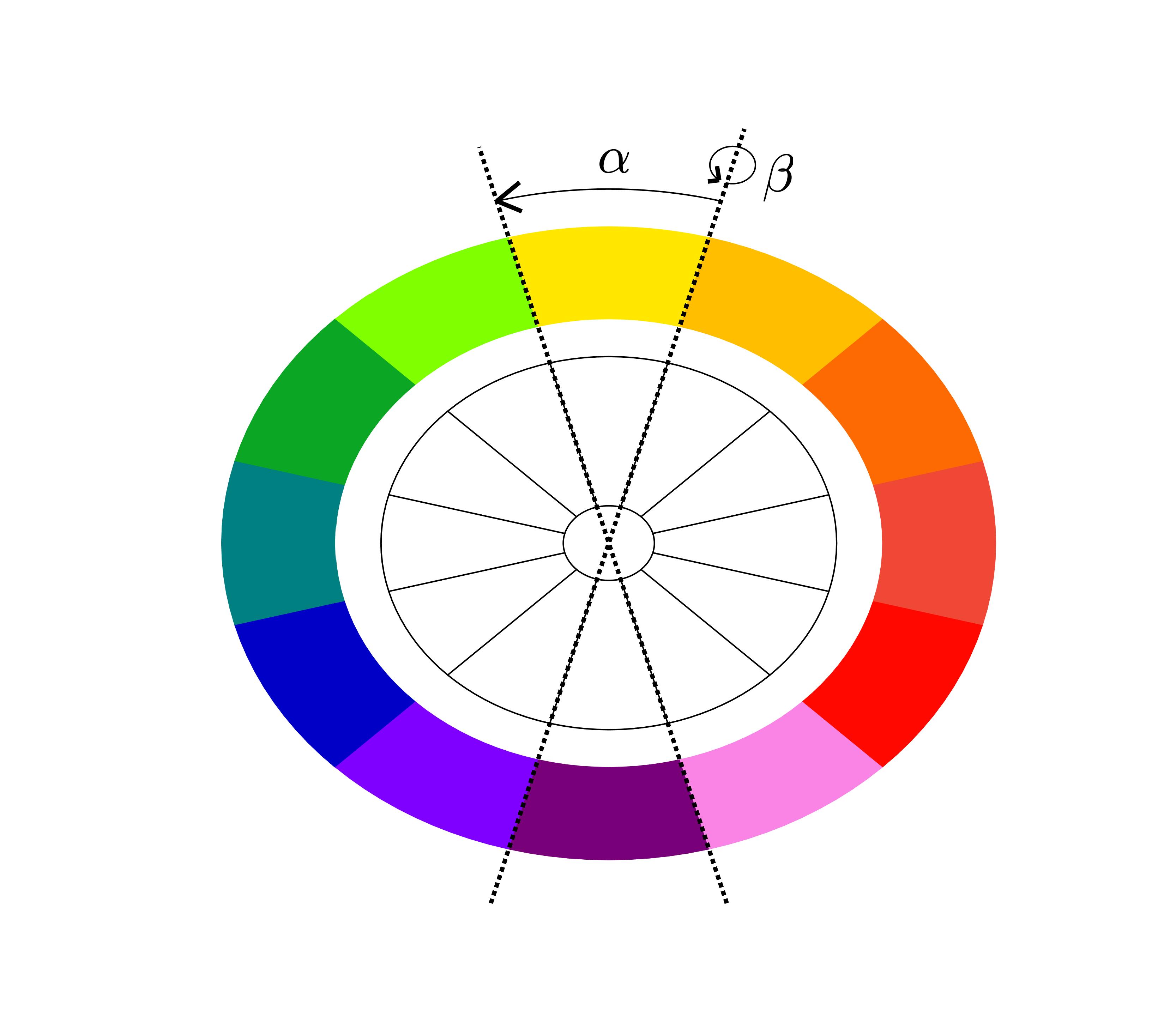}
\includegraphics[trim = 220mm 160mm 180mm 140mm,clip,width=6.8cm, height=7.2cm]{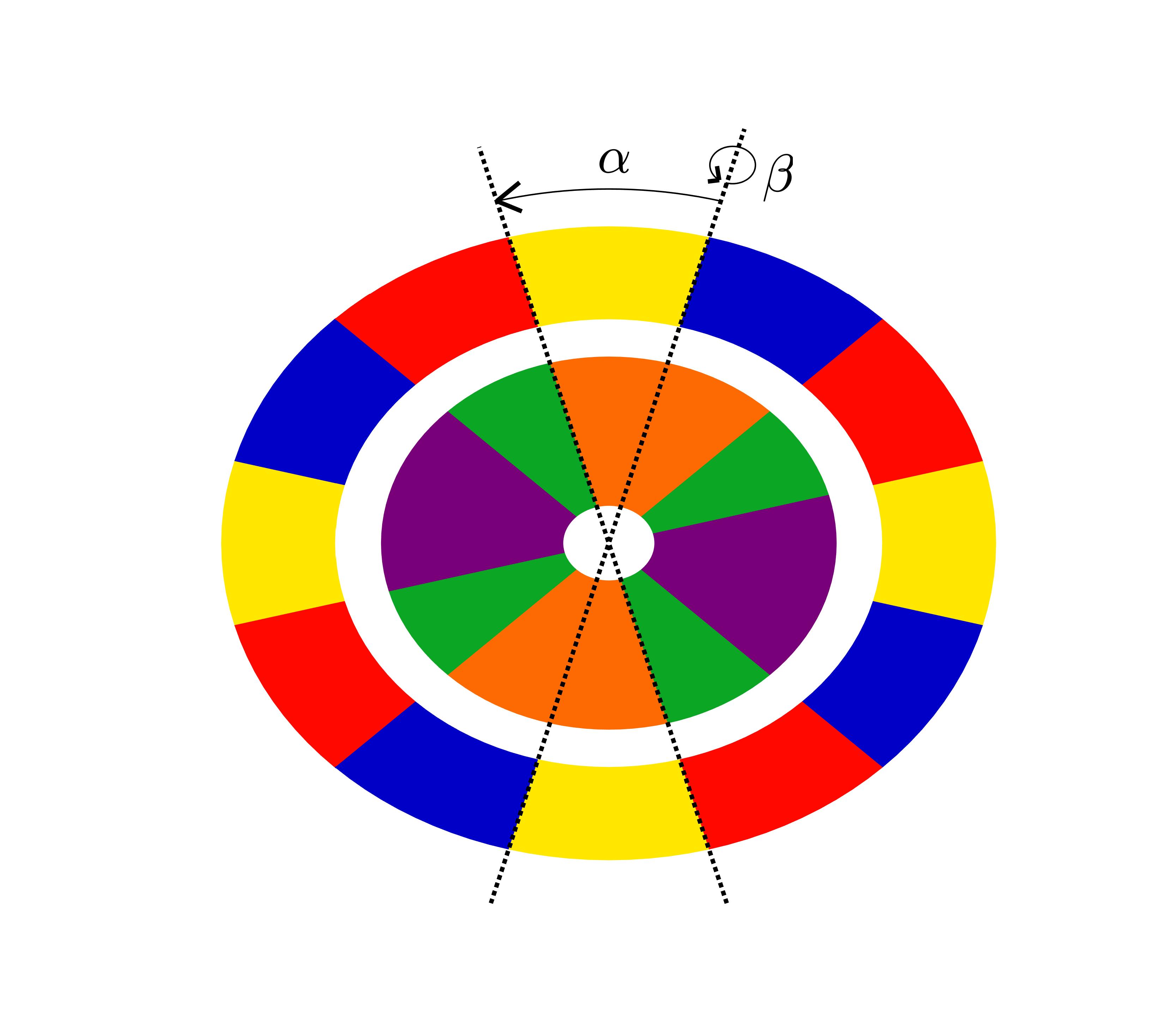}

(a) \hspace{3.0cm} (b)

\caption{Symmetry and color symmetry on a  color wheel with 12 slots. We denote the slots according to the digits of a clock face. Axes connect slots on the right-hand side of each slot. A rotation $\alpha$ rotates counter-clockwise with 30$^\circ$ and a reflection $\beta$ reflects in the axis connecting 12 and 6 o'clock. \textbf{(a)} The outer circle is the RYB (red yellow blue) color wheel. \textbf{(b)} Color symmetry described by the color permutations of Eq. \eqref{eq:col_per1} (outer circle) and Eq. \eqref{eq:col_per2} (inner circle).       
}
\label{fig:color_wheel}
\end{figure}

Suppose we have a circular structure (a wheel) with 12 slots as in Fig. \ref{fig:color_wheel}. Each of the two wheels consists of an inner and an outer circle. Now, let us define a slot to be a motif attributable with a color. Thus, each circle consists of 12 copies of the motif and we can discuss symmetry properties.   In this discussion, we may assign symmetry to the copies of the motif regardless of the color, or taking into account the color. 
The former implies geometrical symmetry described by isometric transformations, the latter leads us to color symmetry. 

First, assume the  color of the slots is not taken into account or the color of all the slots is the same,
see for instance the inner circle of Fig. \ref{fig:color_wheel}a. We denote the slots according to the digits of a clock face with axes connecting slots on the right-hand side of the slot.  Then, the symmetry group of the inner circle is the dihedral group $D_{12}$ (or $\ast 12 \bullet$ in orbifold notation). Let $\alpha$ be a rotation counter-clockwise with 30$^\circ$ and $\beta$ be a reflection in the axis connecting 12 and 6 o'clock, see Fig. \ref{fig:color_wheel}a for illustration. We define that symmetries are composed from left to right. For instance, $\beta \alpha$ is first a reflection $\beta$ and then a rotation $\alpha$. Thus,  $\beta \alpha$ is a reflection in the axis connecting 11 and 5 o'clock. Likewise, $\beta \alpha^2$ is a reflection in the axis connecting 10 and 4 o'clock, and so on.    
The symmetry group $\mathcal{G}$ is generated by $\alpha$ and $\beta$ with $e=\alpha^{12}=\beta^2$, $\beta \alpha \beta=\alpha^{11}$. We obtain \begin{equation*}
    \mathcal{G}=\{e,\alpha,\alpha^2,\ldots,\alpha^{11},\beta,\beta \alpha, \beta \alpha^2,\ldots,\beta \alpha^{11}\},
\end{equation*}  which is of order 24.

We now add color as an attribute to the slots. The 12 slots can be colored with up to 12 different colors, see the outer circle of Fig. \ref{fig:color_wheel}a and both circles of   Fig. \ref{fig:color_wheel}b. The colors of the outer circle of Fig. \ref{fig:color_wheel}a match the   RYB (red-yellow-blue) color wheel~\cite{itten73,rhyne17}, which is also called the standard artistic color wheel. It defines three primary colors: \textbf{r}ed, \textbf{y}ellow and \textbf{b}lue at 4, 12 and 8 o'clock, respectively,  on the circle. Mixing two of these colors each gives the three secondary colors: \textbf{o}range (red and yellow), \textbf{p}urple (red and blue) and \textbf{g}reen (yellow and blue) at 2, 6 and 10 o'clock. From these three primary and three secondary colors, we get another six tertiary colors by mixing: \textbf{ve}rmilion (red and orange), \textbf{am}ber (orange and yellow), \textbf{ch}artreuse (yellow and green), \textbf{te}al (green and blue), \textbf{vi}olet (blue and purple) and \textbf{ma}genta (purple and red).  The colors of the circles of   Fig. \ref{fig:color_wheel}b are also  taken from the RYB color wheel.

Now consider the color symmetry of the colored slots. An element $g \in \mathcal{G}$ is called a
color symmetry if $g$ permutes the coloring of the slots' colors  consistently with the symmetry group $\mathcal{G}$~\cite{roth82,sen83,sen88}. In other words, $\mathcal{G}$ pre-defines the coloring of each copy of the motif. 
More specifically, some (or all) elements in $\mathcal{G}$  change colors, while the remaining elements in 
$\mathcal{G}$ preserve colors. Particularly from an artistic point of view, color symmetry becomes interesting if there is a mix between color-changing and color-preserving elements of $\mathcal{G}$.

The color symmetry of the outer circle of Fig. \ref{fig:color_wheel}a, which is the RYB circle with 12 colors,  is trivial. Of the 24 elements of $\mathcal{G}$, all non-trivial elements change colors, while only the identity $\alpha^{12}=\beta^2=e$ preserves colors. As more interesting color symmetries are obtained if the number of copies of the motif exceeds the number of colors, we
next consider the color circles  with 3 colors, see  Fig. \ref{fig:color_wheel}b, where the outer circle takes the primary colors and the 
 inner circle takes the  secondary  colors.  
 For such a threefold coloration, the color permutation has order $3$. Recall that the symmetry group $\mathcal{G}$ of the 12-slot circle specifies counter-clockwise rotations about 30$^\circ$ and reflections along 12 axis. For the outer circle, all reflections $\{\beta,\beta \alpha,\beta \alpha^2,\ldots,\beta \alpha^{11}\}$ change colors (except the trivial $\beta^2=e$).  Rotations $\alpha$ change or preserve the colors according to a color permutation. Using Cauchy's two-line notation, the color permutation associated with the rotation $\alpha$ can be expressed by \begin{equation}
     p_{\alpha|\alpha^2|\alpha^3}=\left(\begin{smallmatrix} \mathbf{r} & \mathbf{y}& \mathbf{b} \\ \mathbf{b} & \mathbf{r} & \mathbf{y}  \end{smallmatrix} \right| \left. \begin{smallmatrix} \mathbf{r} & \mathbf{y}& \mathbf{b} \\ \mathbf{y} & \mathbf{b} & \mathbf{r}  \end{smallmatrix} \right| \left. \begin{smallmatrix} \mathbf{r} & \mathbf{y}& \mathbf{b} \\ \mathbf{r} & \mathbf{y} & \mathbf{b}  \end{smallmatrix} \right) \label{eq:col_per1} .
 \end{equation}   The color permutation implies that the colors are preserved for the rotations $\alpha^3$, $\alpha^6$ and $\alpha^9$, which are the rotations about 90$^\circ$.
 
 For the inner circle, almost all rotations $\{\alpha,\alpha^2,\ldots,\alpha^{11}\}$ change color (except the trivial $\alpha^{12}=e$ and $\alpha^{6}$). Reflections $\beta$ change or preserve the colors according to the color permutation \begin{equation}
     p_{\beta|\beta\alpha|\beta\alpha^2}=\left(\begin{smallmatrix} \mathbf{o} & \mathbf{g}& \mathbf{p} \\ \mathbf{o} & \mathbf{g} & \mathbf{p}  \end{smallmatrix} \right| \left.  \begin{smallmatrix} \mathbf{o} & \mathbf{o}& \mathbf{g} \\ \mathbf{g} & \mathbf{p} & \mathbf{p}  \end{smallmatrix} \right| \left.  \begin{smallmatrix} \mathbf{g} & \mathbf{o}& \mathbf{o} \\ \mathbf{p} & \mathbf{p} & \mathbf{g}  \end{smallmatrix}  \right), \label{eq:col_per2}
 \end{equation}  which implies that the reflections $\beta,\beta \alpha^3,\beta \alpha^6$ and $\beta \alpha^9$ preserve colors. 
 
 The color symmetry on a wheel with a finite number of slots such as in Fig.  \ref{fig:color_wheel} implies a transitive color permutation, as shown with the examples   $p_{\alpha|\alpha^2|\alpha^3}$ and $p_{\beta|\beta\alpha|\beta\alpha^2}$, see Eqs. \eqref{eq:col_per1} and \eqref{eq:col_per2}. In other words, for any pair of colors on the wheel, there is an element $g \in \mathcal{G}$ which maps the one color of the pair to the other~\cite{roth82}. Possible generalizations could mean that one color is ``color-symmetric'' to two or more colors, or that an element   $g \in \mathcal{G}$ is connected with more than one color mapping, for instance by applying to different sections of the motif differently.  Another implication of the color symmetry discussed so far is that it applies to motifs which are either pre-existing or designed somehow else.  We next discuss color symmetry which  applies to pre-existing images.  Suppose we have a rectangular source images, for instance the Kandinsky study shown in Fig. \ref{fig:kandins}a. The symmetry group of the rectangle is the dihedral group $D_2$. The rotation $\alpha$ is now counter-clockwise  180$^\circ$, the reflections $\beta_h$ and $\beta_v$ are in the horizontal and vertical axis. The symmetry group is of order 4: \begin{equation*}
     \mathcal{G}=(e,\alpha,\beta_h, \beta_v).
 \end{equation*}  The reflection axes divide the image into 4 sections, north-east (\textbf{ne}), north-west (\textbf{nw}), south-east (\textbf{se}) and south-west (\textbf{sw}). Each section can be mapped into any other section by the elements of the symmetry group. The sections are separated in    
Fig. \ref{fig:kandins}a by white lines (which are not present in the original Kandinsky study).  We assume that the image is composed of $n \times m$ pixels and each pixel $p_i$ has a color $c_i$ in a HSV color space: $c_i=(h_i,s_i,v_i)$. In the following discussion (and also in the experiments reported later) the focus is on the hue $h_i$ while the saturation $s_i$ and the value (brightness) $v_i$ remain constant, but of course the same method can be applied to the whole HSV color space. 
We now define the color permutation associated with the symmetry group by a map $f(h_i)$, which is a mapping of the hue component of the HSV color space onto itself.  Thus, for rectangular source images, a color symmetry  
can be imposed by a color permutation as follows. We define a reference section, for instance  north-east (\textbf{ne}). The colors in this section remain unchanged. The colors of the pixels of the other sections are changed according  to the rectangle's symmetry group. For instance, the color permutation associated with the rotation is \begin{equation}
    p_\alpha=\left(\begin{smallmatrix} \mathbf{h_i}  \\ \mathbf{f_1(h_i)}   \end{smallmatrix} \right), \label{eq:col_per11}
\end{equation} which implies that the hue $h_i$ of all pixels $p_i$ in
\textbf{sw} is changed from $h_i$ to $f_1(h_i)$. For the remaining sections \textbf{nw} and \textbf{se}, we may likewise define for the reflections \begin{equation}
    p_{\beta_v}=\left(\begin{smallmatrix} \mathbf{h_i}  \\ \mathbf{f_2(h_i)}   \end{smallmatrix} \right),  \qquad p_{\beta_h}=\left(\begin{smallmatrix} \mathbf{h_i}  \\ \mathbf{f_3(h_i)}   \end{smallmatrix} \right). \label{eq:col_per21}
\end{equation} 
Fig. \ref{fig:kandins}b shows an example using the Kandinsky study (Fig. \ref{fig:kandins}a) as a source image. As color maps the functions $f_1(h_i)=0.45 |\sin{(\sqrt{2}\cdot 20 \pi h_i})|+0.55 |\sin{(20 \pi h_i)}|$, $f_2(h_i)=0.5(1+\sin{(40 \pi h_i)})$ and $h_3(h_i)=4h_i(1-h_i)$ are used.

Such a design of color symmetry implies that the color permutations according to Eqs. \eqref{eq:col_per11} and \eqref{eq:col_per21} most likely still have finite order.
Of course, any color described by the hue $h_i \in \mathbb{R}$ could be mapped to another color $f(h_i)$. 
But most likely the HSV color space used in computer graphics effectively imposed a finite number of different colors. 
Nevertheless, the number of different colors surely is very large, which opens up a rich variety of color effects. The visual result in Fig.  \ref{fig:kandins}b shows a substantial variety of color shifts in the sections affected by color symmetry, particularly for the color maps involving trigonometric functions. Moreover, 
as the saturation and brightness of the pixels are preserved while the hue is changed, the ``color symmetric'' sections of the image resembles the original image, but we also have a kind of alienation effect (or distancing effect) in terms of color.

 In Sec. \ref{sec:exp} generative art is presented and discussed which is the result of experiments with color, symmetry, and color symmetry. This is done in connection with stigmergic generation algorithms, which are discussed next.

\section{Stigmergy and generative art} \label{sec:stig}
Stigmergy is a bio-inspired paradigm of self-organisation in swarms. The members of the swarm coordinate themselves by indirect communication, for instance by deposing and collecting visual, olfactory or 	
tactile markers. This paradigm has been used as a computational framework for generating swarm art, see for examples~\cite{al13,fernand14,green12,green14,green15,jac07,neu20,rich18}.

In the experiments with designing color symmetry presented in this paper we use a model of stigmergic nest construction. The model describes the nest building by the ant species {\it T. albipennis} and was proposed by Urbano~\cite{urb11}. 
The nests are constructed by the ants collecting sand grains from a foraging ground and deposing them to build a closed curved outer boundary wall around the nest. Different colonies of ants may be sensitive to sand with different colors, thus making it possible to construct polychromatic swarm paintings. 
The nest construction modelling {\it T. albipennis} behavior  builds toroidal walls~\cite{green12,urb11}. However, it was suggested by Greenfield~\cite{green14} that the boundary walls could also follow other types of curves, which would make them more interesting  than circles from a geometrical point of view. 
In such an interpretation of boundary walls, different colonies would be sensitive to different types of walls.
In the following, this idea is adopted and the stigmergic art discussed in this paper is constructed from the following algorithmic framework. Each nest has a center point $P_i=(x_i,y_i)$ in a 2D plane. The nest is further characterized by a boundary wall with radius $r_i(\theta)$ which describes the distance to the center point in polar coordinates with $\theta=[0,2k\pi]$, $k=1,2,3,\ldots$. Thus, by setting $k=1$, circles, for instance, can be constructed by $r_i(\theta)=r_i=const.$,  and $\ell$-leaf roses are obtained for $r_i(\theta)=a \sin(\ell \theta)$ and $r_i(\theta)=a \cos(\ell \theta)$.  
Archimedean spirals are described by $r_i(\theta)=\gamma+\delta \theta$, with $\gamma,\delta>0$ and $k$ specifying the desired length of the spiral.

The nest's boundary wall is build by ants collecting grain with the nest's boundary type from the foraging ground. An ant has a certain distance from the nest's center point and can either carry a grain or not.    If it carries a grain it moves randomly toward the center point and deposes the sand grain with a certain probability within the vicinity of the radius $r_i(\theta)$. By parameters we can control the speed of nest formation and also the wall thickness. If the ant does not carry a grain it moves randomly away from the center point and   
collects a sand grain with a second probability. Again, we can control by parameters the collecting speed and foraging range. The visual effect is that boundaries walls are built which have a grain density resembling realizations of a random variable with centre    $r_i(\theta)$, for instance a Gaussian distribution with $\mathcal{N}(r_i(\theta),\sigma)$. Contrary to previous works~\cite{green12,green14,urb11}, we assume that the  foraging ground is completely emptied in the nest building activity. 
Therefore, we can describe the $i$-th nest by its center point, its color and its radius: $N_i=(x_i.y_i,c_i,r_i(\theta))$.
We consider these nests as motifs.

\section{Experiments and visual results} \label{sec:exp}

  \begin{figure}[htb]
 
\centering
\includegraphics[trim = 80mm 60mm 70mm 30mm,clip,width=7.35cm, height=6.15cm]{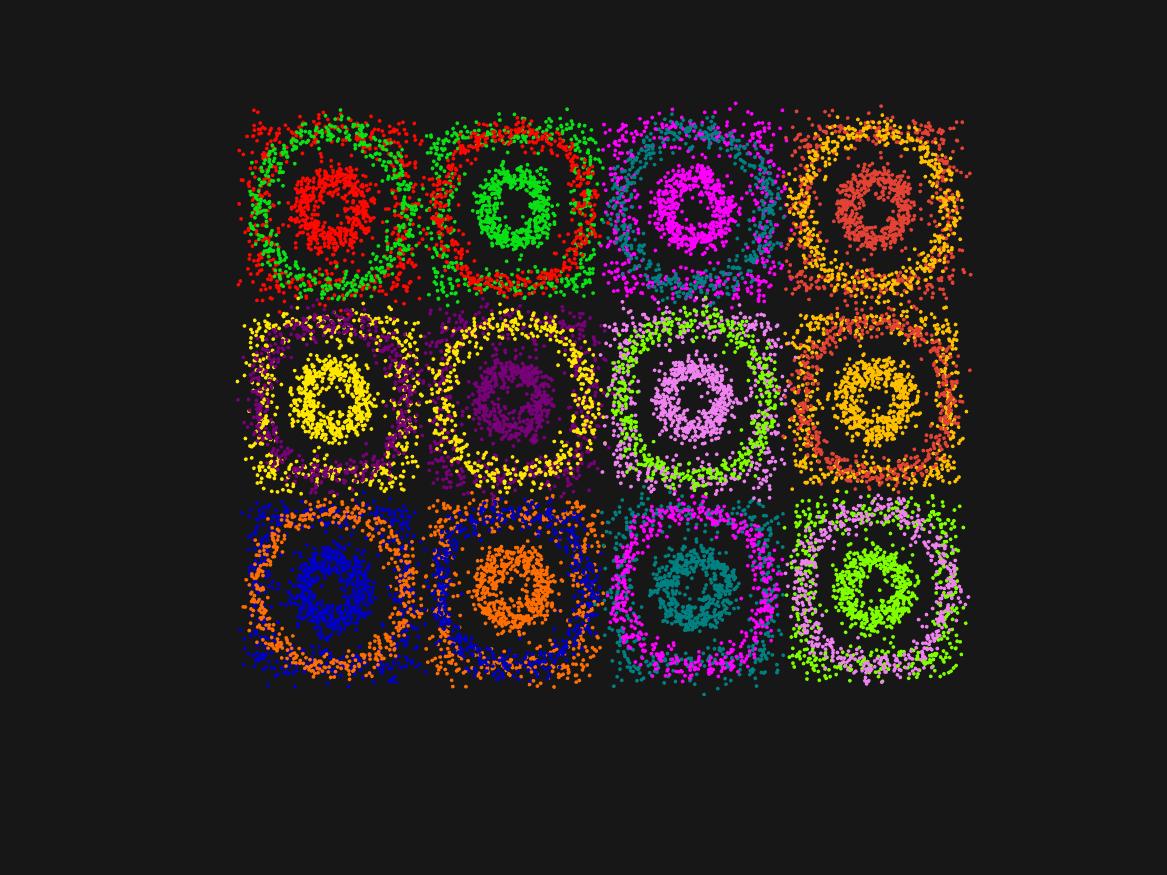}
\includegraphics[trim = 80mm 60mm 70mm 30mm,clip,width=7.35cm, height=6.15cm]{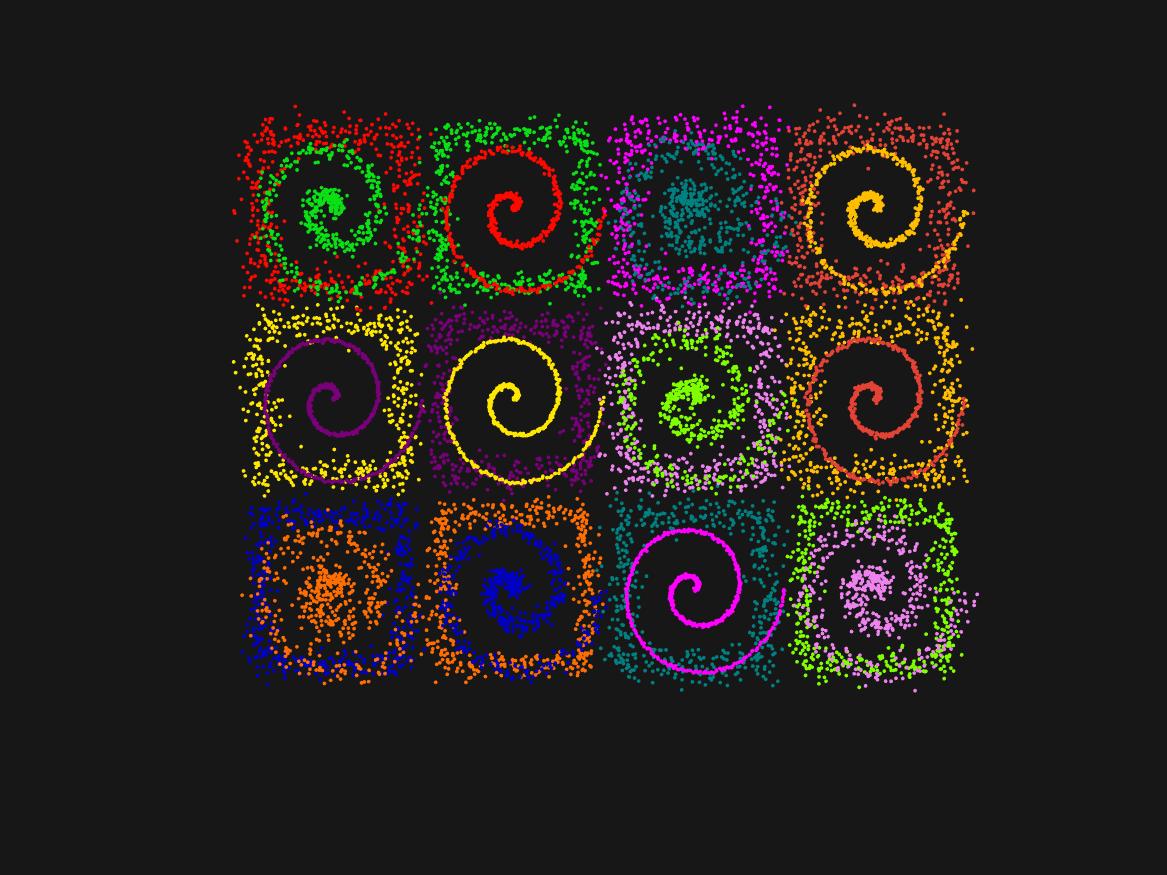}

\vspace{0.1cm}

\includegraphics[trim = 80mm 60mm 70mm 30mm,clip,width=7.35cm, height=6.15cm]{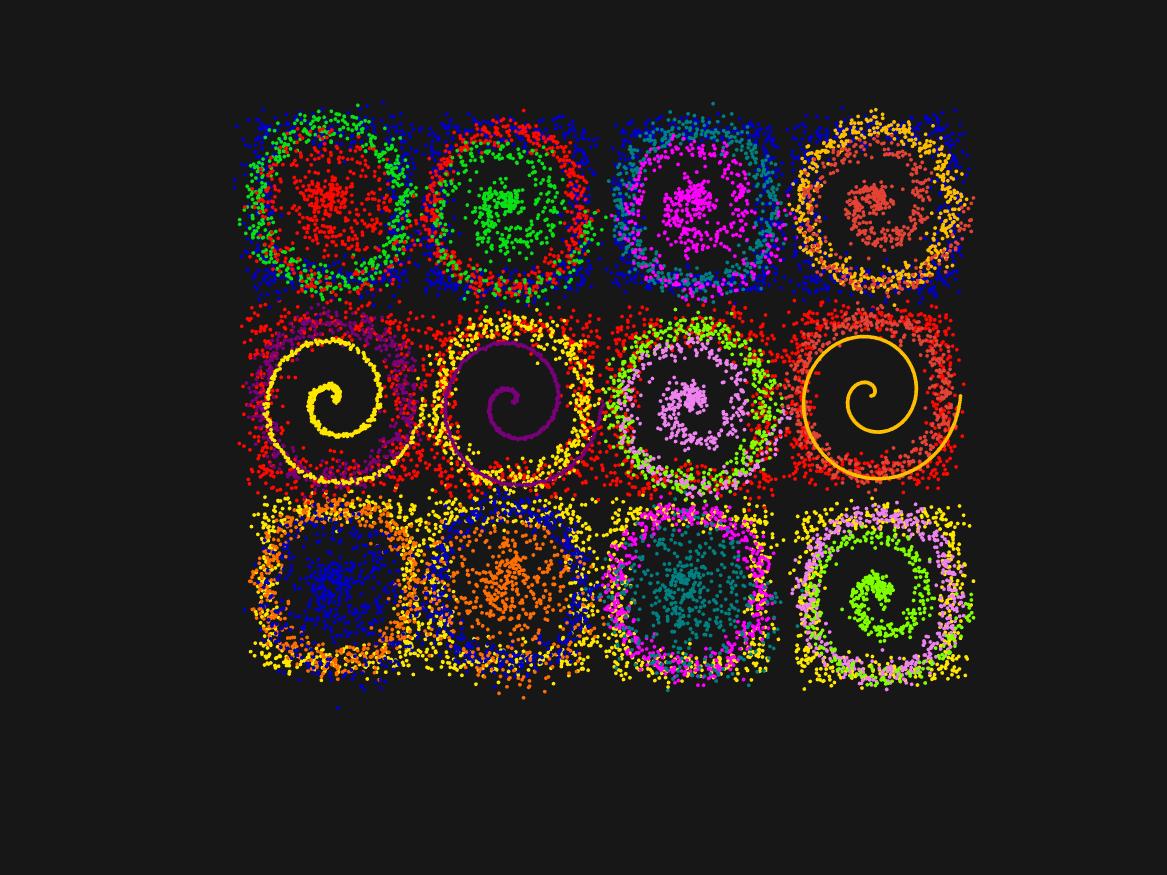}
\includegraphics[trim = 80mm 60mm 70mm 30mm,clip,width=7.35cm, height=6.15cm]{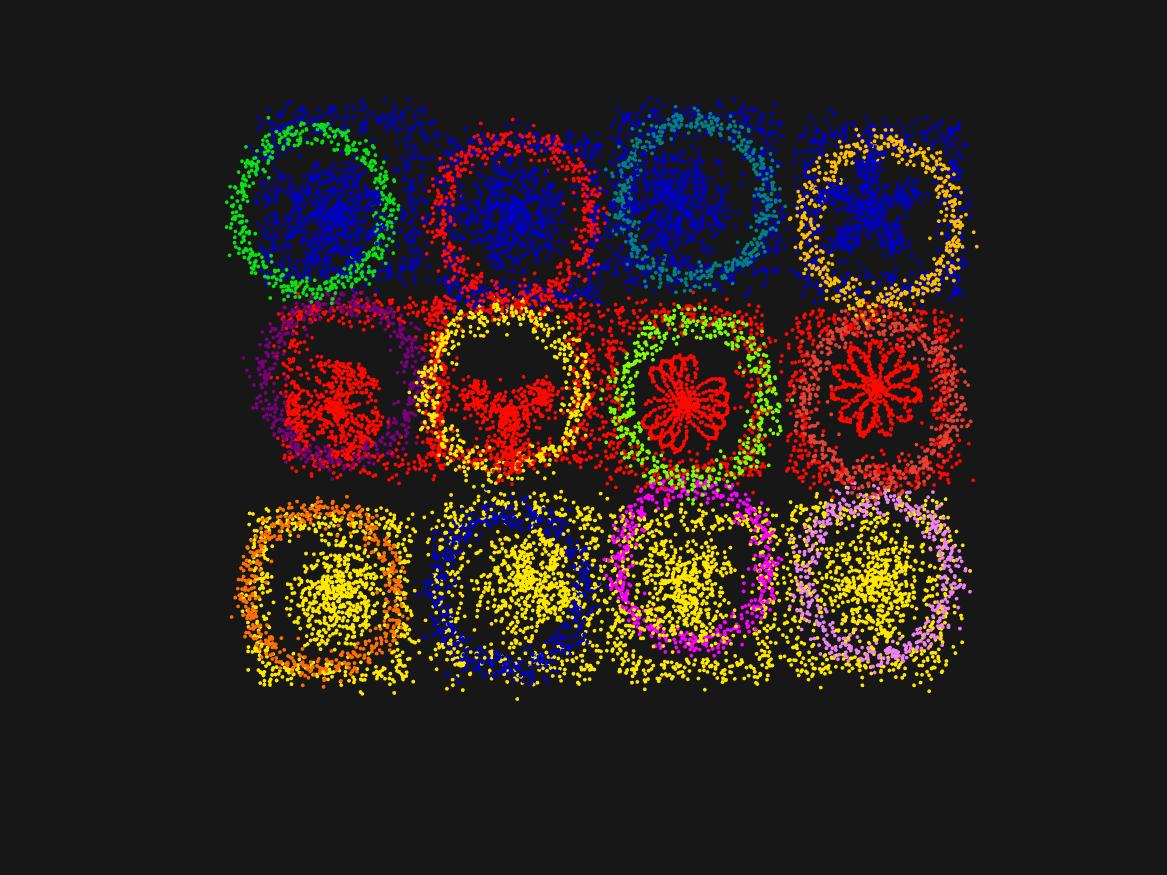}

\caption{Stigmergic color-symmetric patterns with reminiscence about Kandinsky's color study in Fig. \ref{fig:kandins}a.  
}
\label{fig:kand_col_stud}
\end{figure}

  \begin{figure}[htb]
 
\centering
\includegraphics[trim = 80mm 50mm 65mm 20mm,clip,width=7.35cm, height=6.15cm]{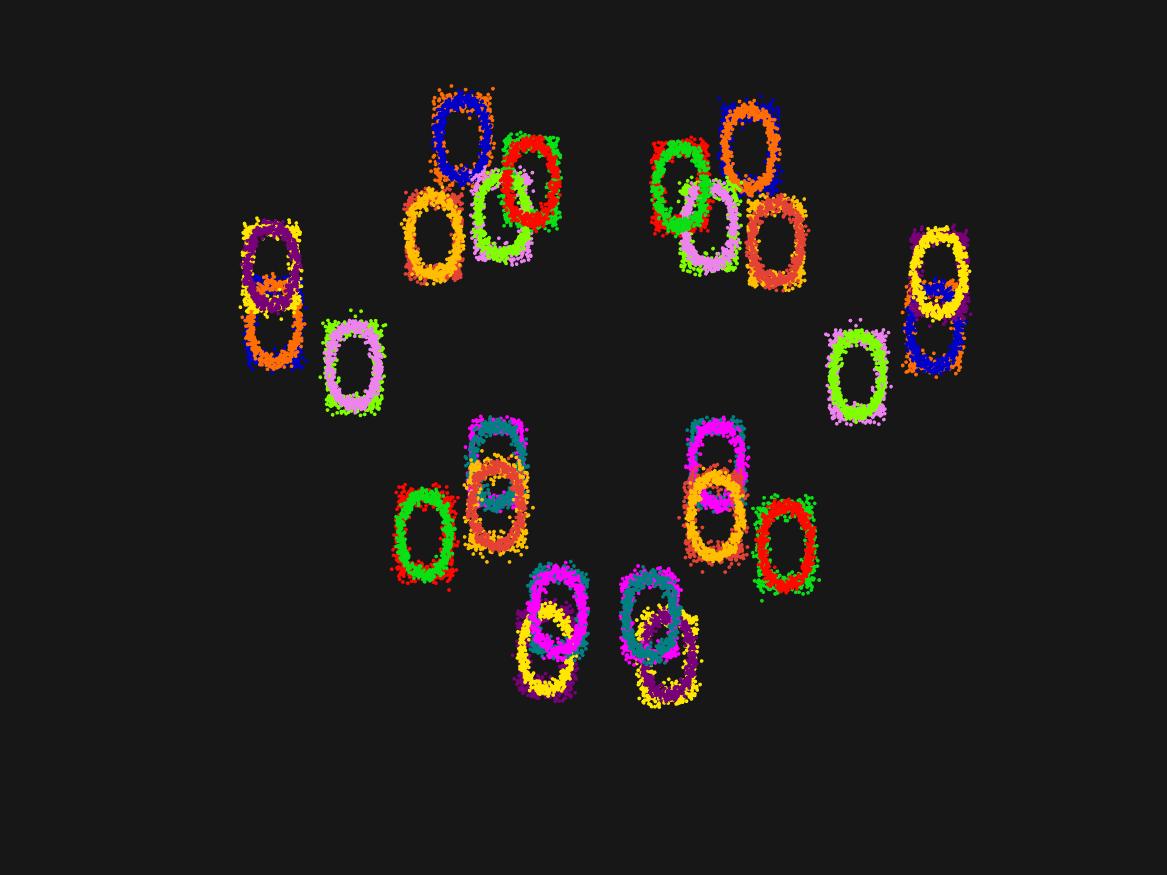}
\includegraphics[trim = 80mm 50mm 50mm 20mm,clip,width=7.35cm, height=6.15cm]{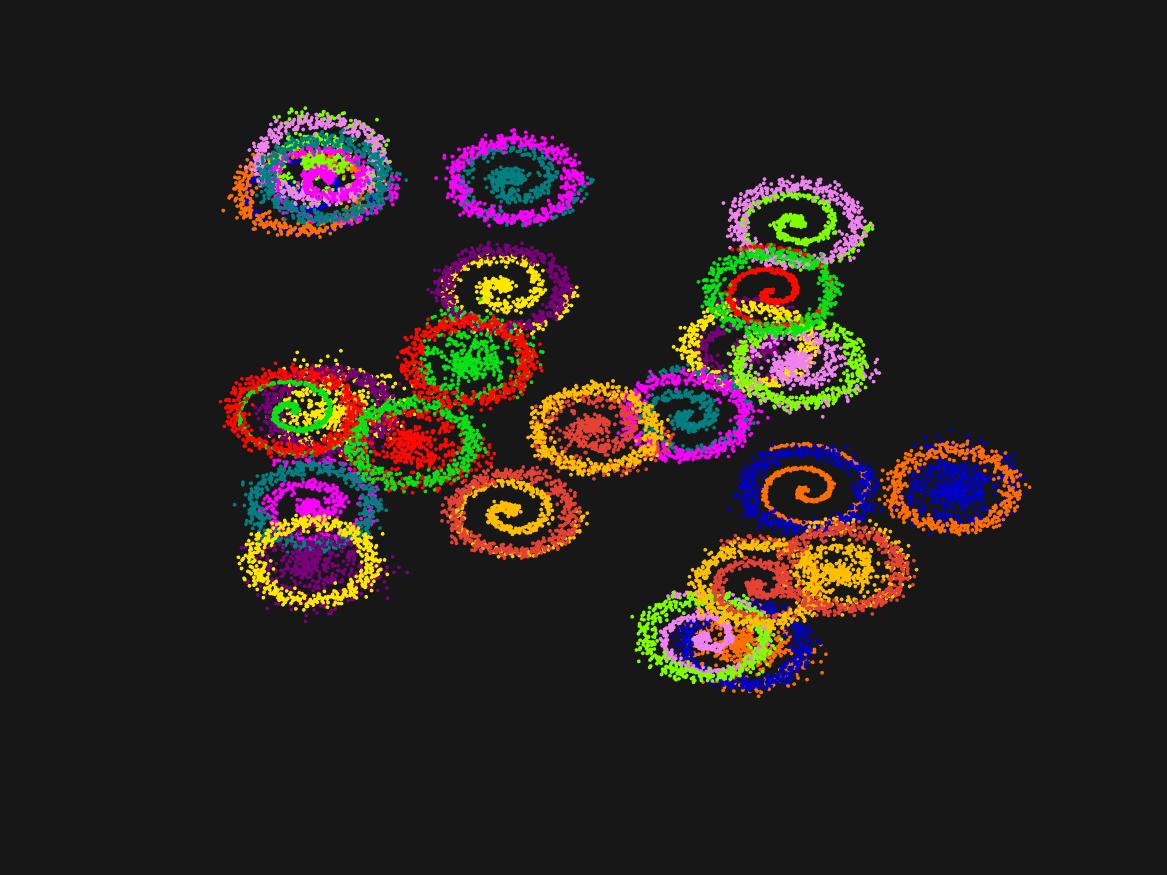}

\vspace{0.1cm}

\includegraphics[trim = 70mm 40mm 60mm 30mm,clip,width=7.35cm, height=6.15cm]{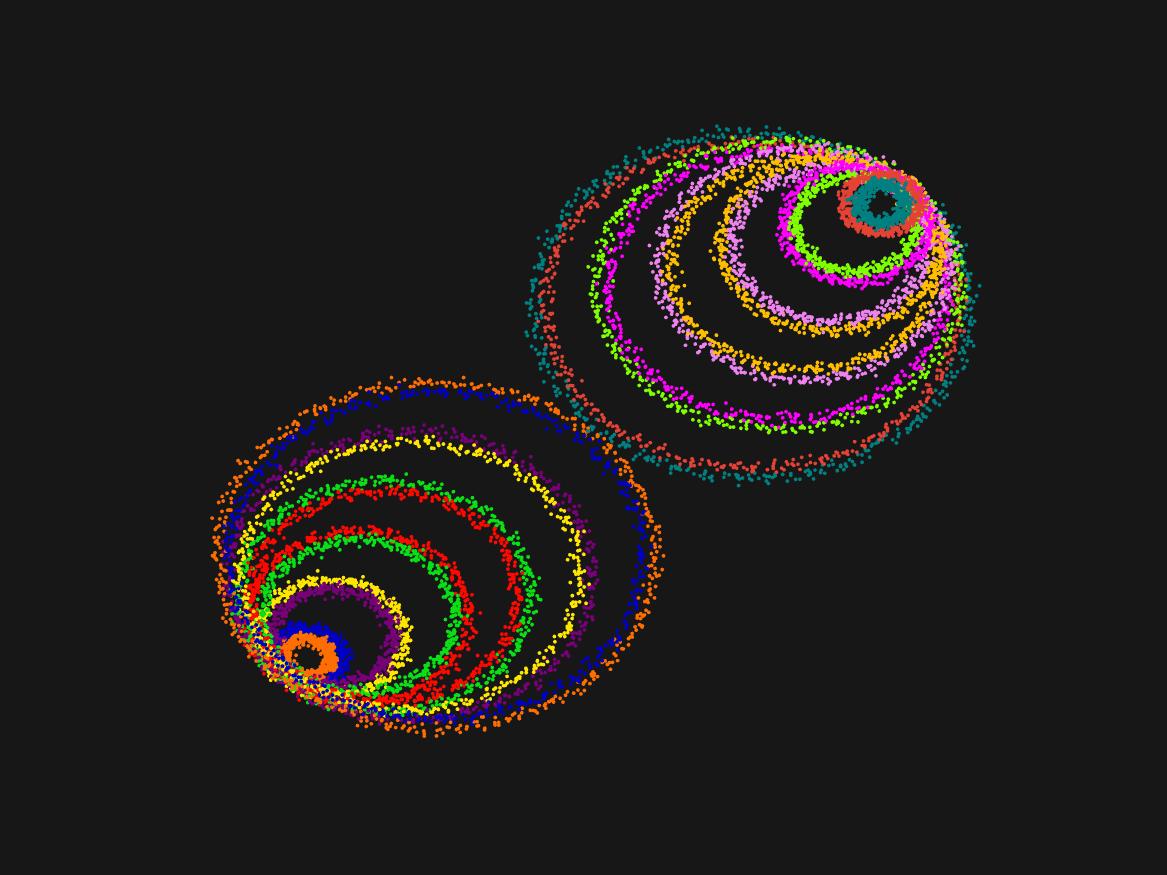}
\includegraphics[trim = 80mm 50mm 50mm 20mm,clip,width=7.35cm, height=6.15cm]{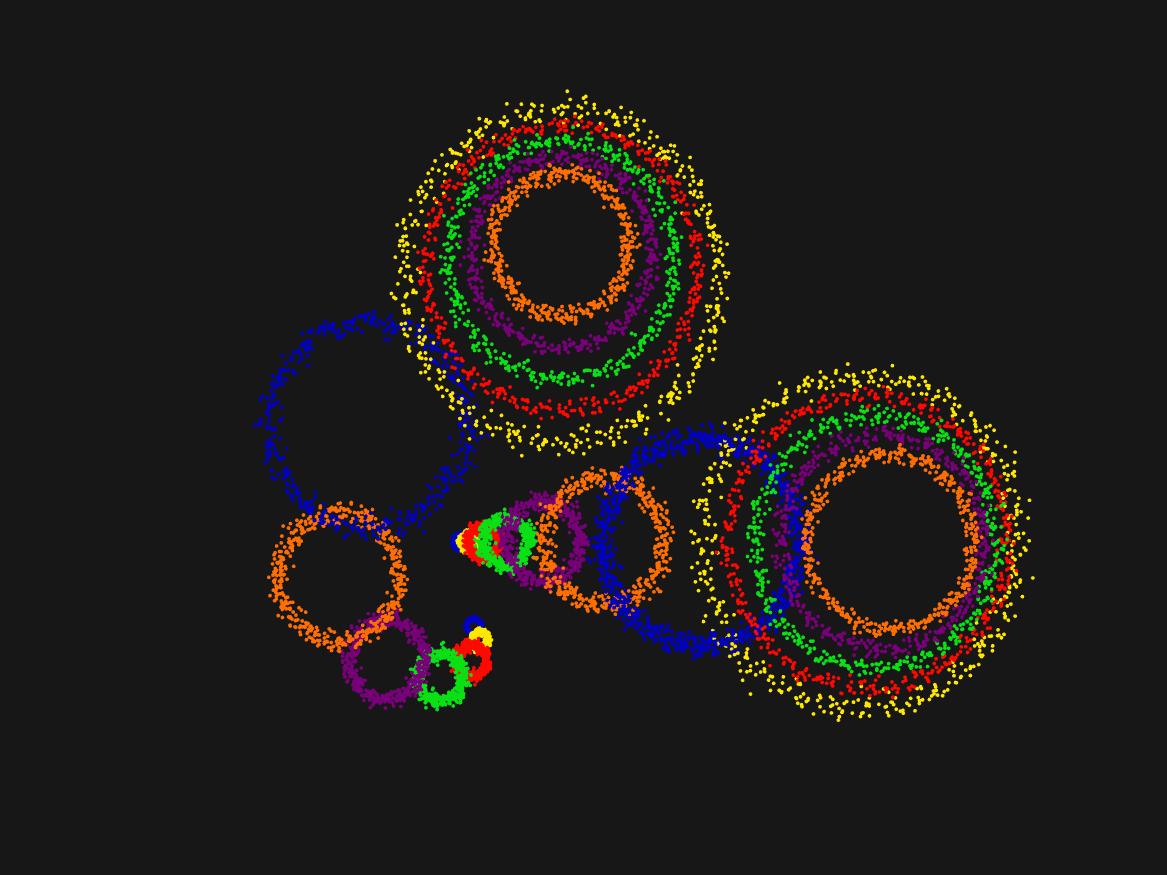}

\caption{Stigmergic color-symmetric patterns  with isometric and color symmetry.
}
\label{fig:kand_spac_stud}
\end{figure}

 \begin{figure}[tb]
 \centering

\includegraphics[trim = 80mm 40mm 70mm 0mm,clip,width=5.4cm, height=5.8cm]{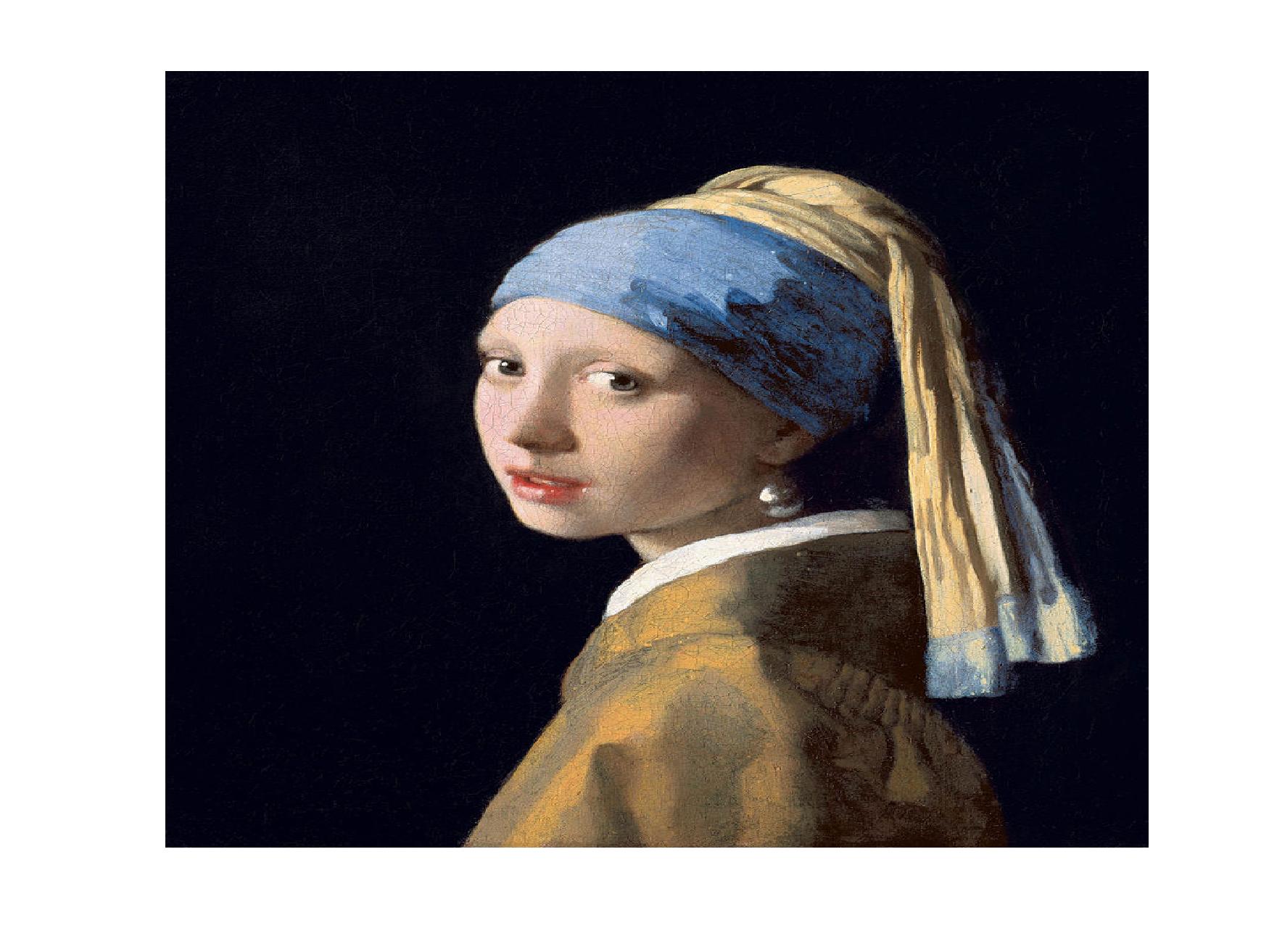}
\includegraphics[trim = 80mm 40mm 70mm 0mm,clip,width=7.2cm, height=5.8cm]{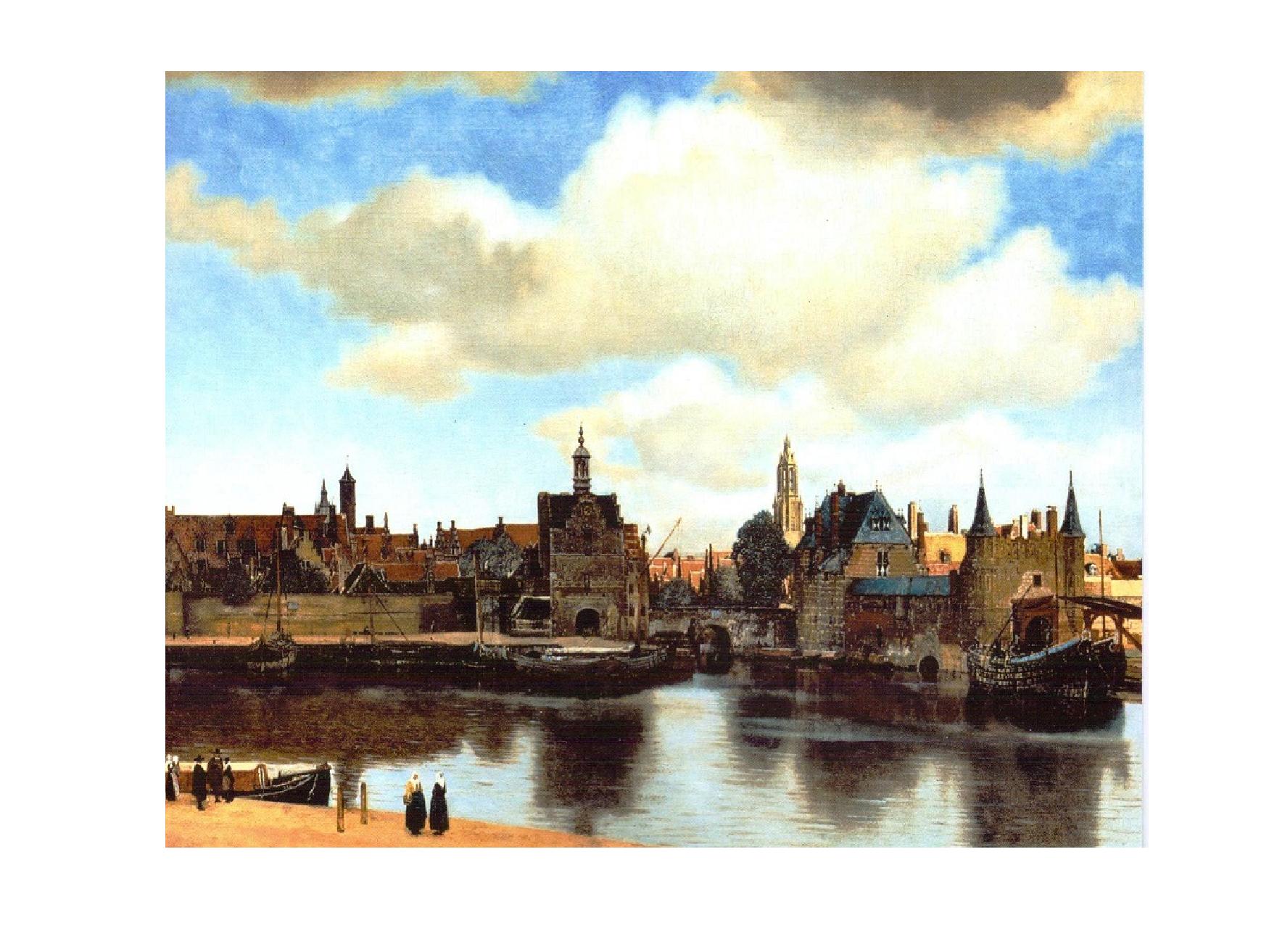}

\hspace{2.1cm}(a) \hspace{3.5cm} (b)

\includegraphics[trim = 80mm 40mm 70mm 0mm,clip,width=5.4cm, height=5.8cm]{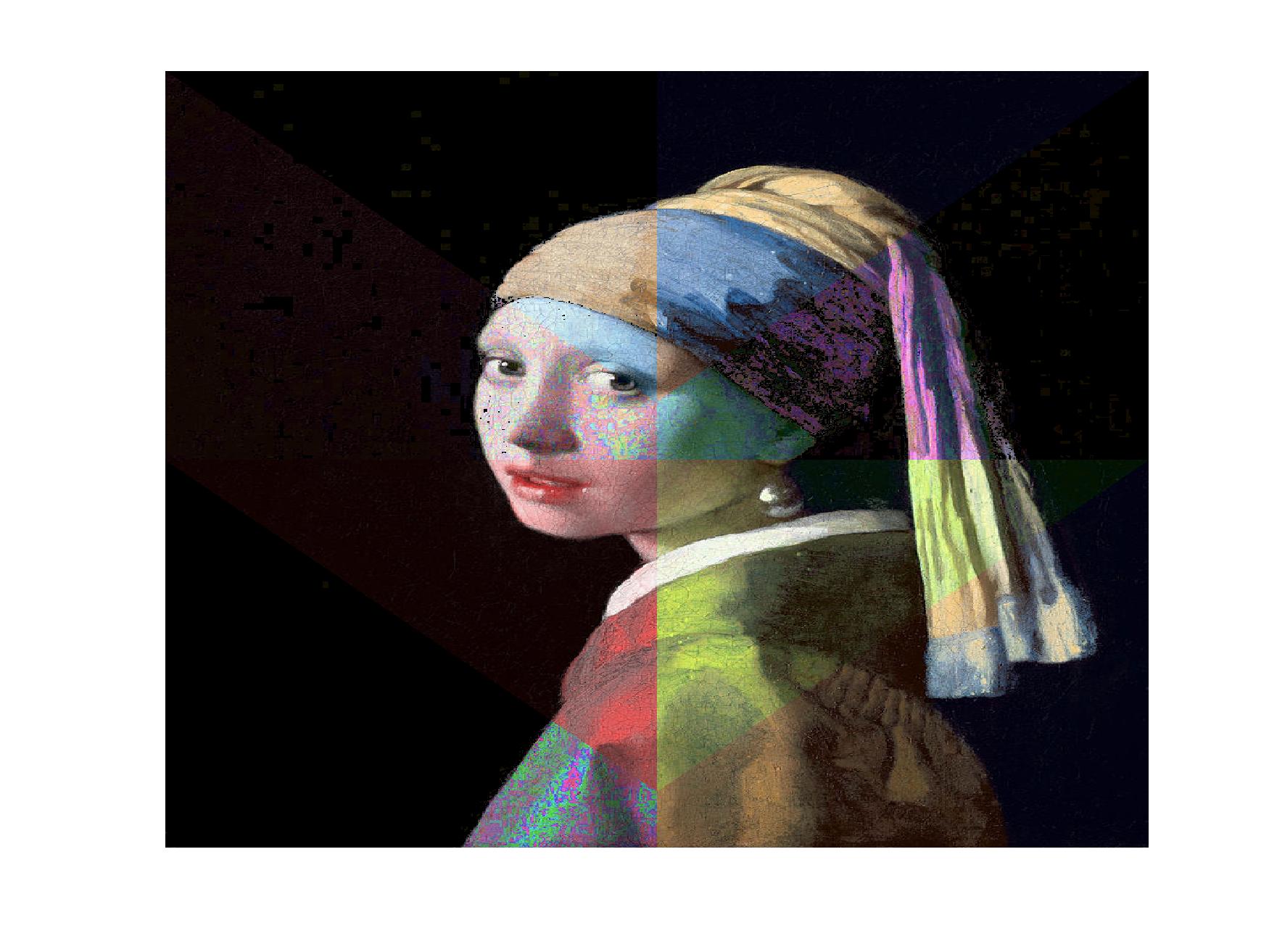}
\includegraphics[trim = 80mm 40mm 70mm 0mm,clip,width=7.2cm, height=5.8cm]{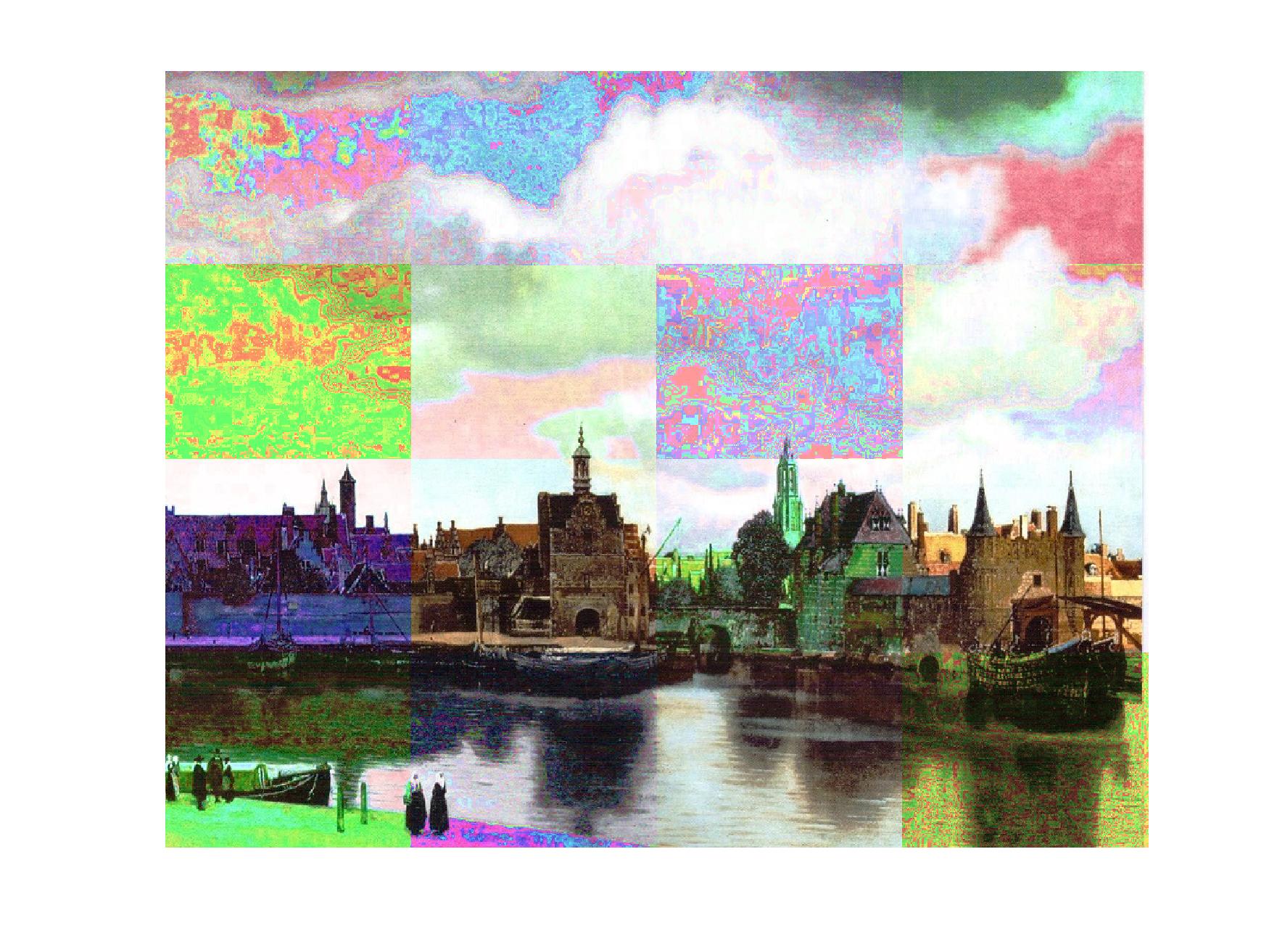}

\hspace{2.1cm}(c) \hspace{3.5cm} (d)

\caption{Johannes Vermeer (1632-1675): \textbf{(a)} {\it Girl with a Pearl Earring},
1665. \textbf{(b)}
{\it View of Delft}, 1661.  The same Vermeer painting with color symmetry imposed by a color permutation. \textbf{(c)} Triangular pattern.  \textbf{(d)} Chess board pattern with 16 elements. 
}
\label{fig:vermeer}
\end{figure}

In the sections above, a framework is set out for stigmergic generation of motifs. In addition, the motifs may be the object of experiments with designing color symmetries.  In the following we present and discuss two sets of artistic experiments generating visual art by this framework. 

A first set of experiments takes the nests generated by stigmergic design, colors the grains according to color-symmetric principles and places them on a black canvas. The examples in Fig.  \ref{fig:kand_col_stud} are directly inspired by  Kandinsky's color study shown in Fig.  \ref{fig:kandins}a. The images seize the idea of 12 squares filling a rectangle in 3 rows. The squares contain circles, but also Archimedean spirals and  $\ell$-leaf roses. The nests are colored with elements from the RYB color wheel (Fig.  
\ref{fig:color_wheel}a). The 12 squares fit exactly the whole set of primary, secondary and tertiary colors of the RYB wheel. Accordingly, the upper left image, for instance, has the 6 squares and the 6 inner circles on the left hand side with primary and secondary colors, while the outer circles are the complementary colors (which are exactly opposite on the RYB color wheel).  The 6 remaining squares on right hand side of the image do the same with the tertiary colors. Thus, a vertical reflection imposes a color symmetry from primary and secondary colors to tertiary colors.  The other images similarly play with squares, circles, Archimedean spirals and  $\ell$-leaf roses and color symmetries between the RYB colors. Particularly note the lower left image where the primary colors are preserved by vertical reflection while secondary colors may change.
The images in Fig. \ref{fig:kand_spac_stud} leave behind the rows-and-columns arrangement and place the nests at any prescribed locations on the canvas. Again, isometric symmetry induces color symmetry. In the upper left image of   Fig. \ref{fig:kand_spac_stud} the geometrical aspect of symmetry is almost invisible, but we have color symmetry insofar as  any circle with an Archimedean spiral has an color-symmetric counterpart which is obtained by a rotation about a random center point. The radii of smallest 7 circles in the lower left image of   Fig. \ref{fig:kand_spac_stud} follow a Fibonacci series with the circles on the left-hand side of the image following a Fibonacci circle curve as proposed by Happersett~\cite{happer14}. The Fibonacci circle curve  places subsequent circles according to the Golden Angle 180$^\circ(3-\sqrt{5})$.

\begin{figure}[htb]
 
\centering
\includegraphics[trim = 100mm 80mm 70mm 20mm,clip,width=5.75cm, height=5.75cm]{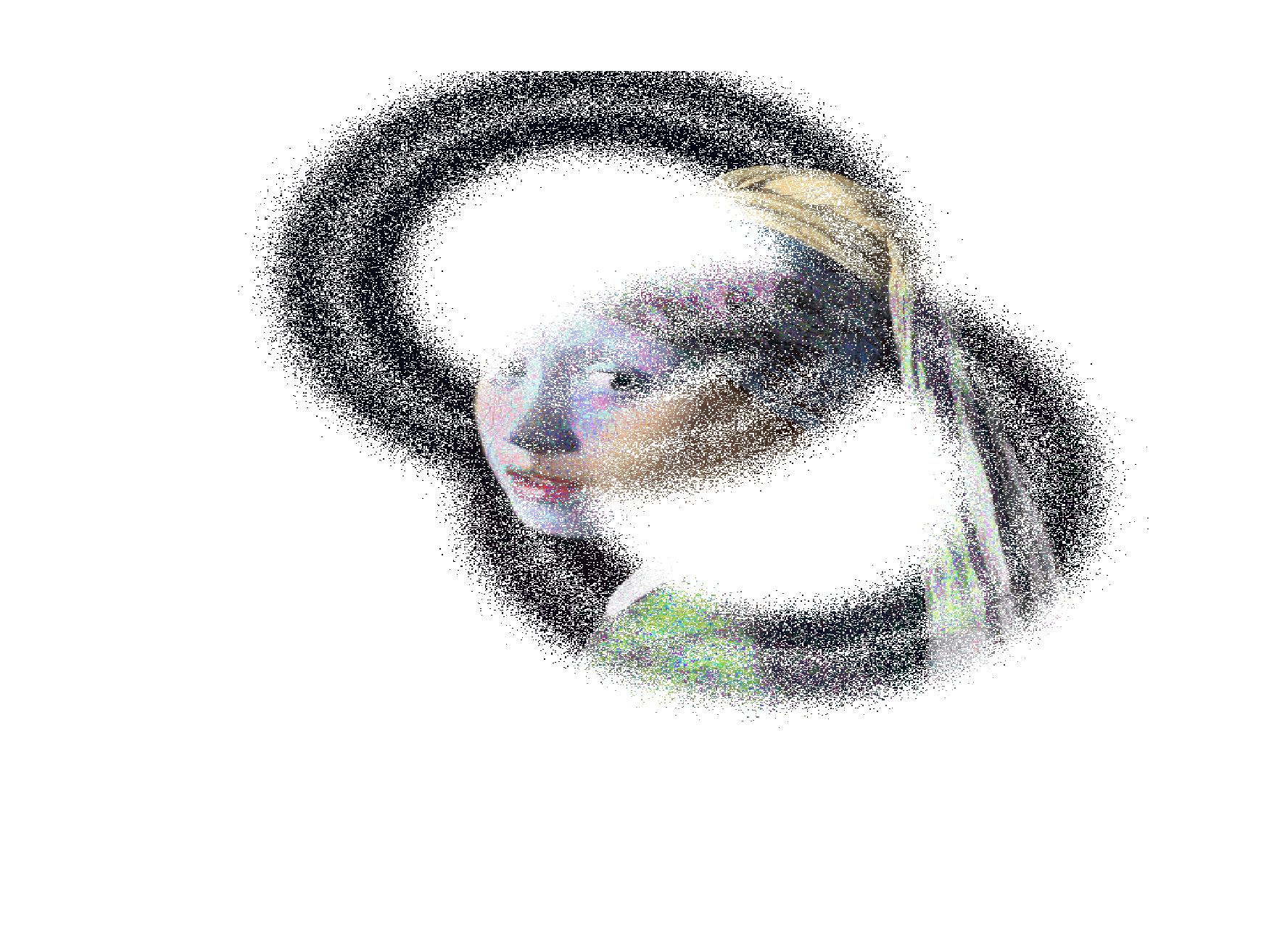}
\includegraphics[trim = 80mm 80mm 70mm 20mm,clip,width=5.75cm, height=5.75cm]{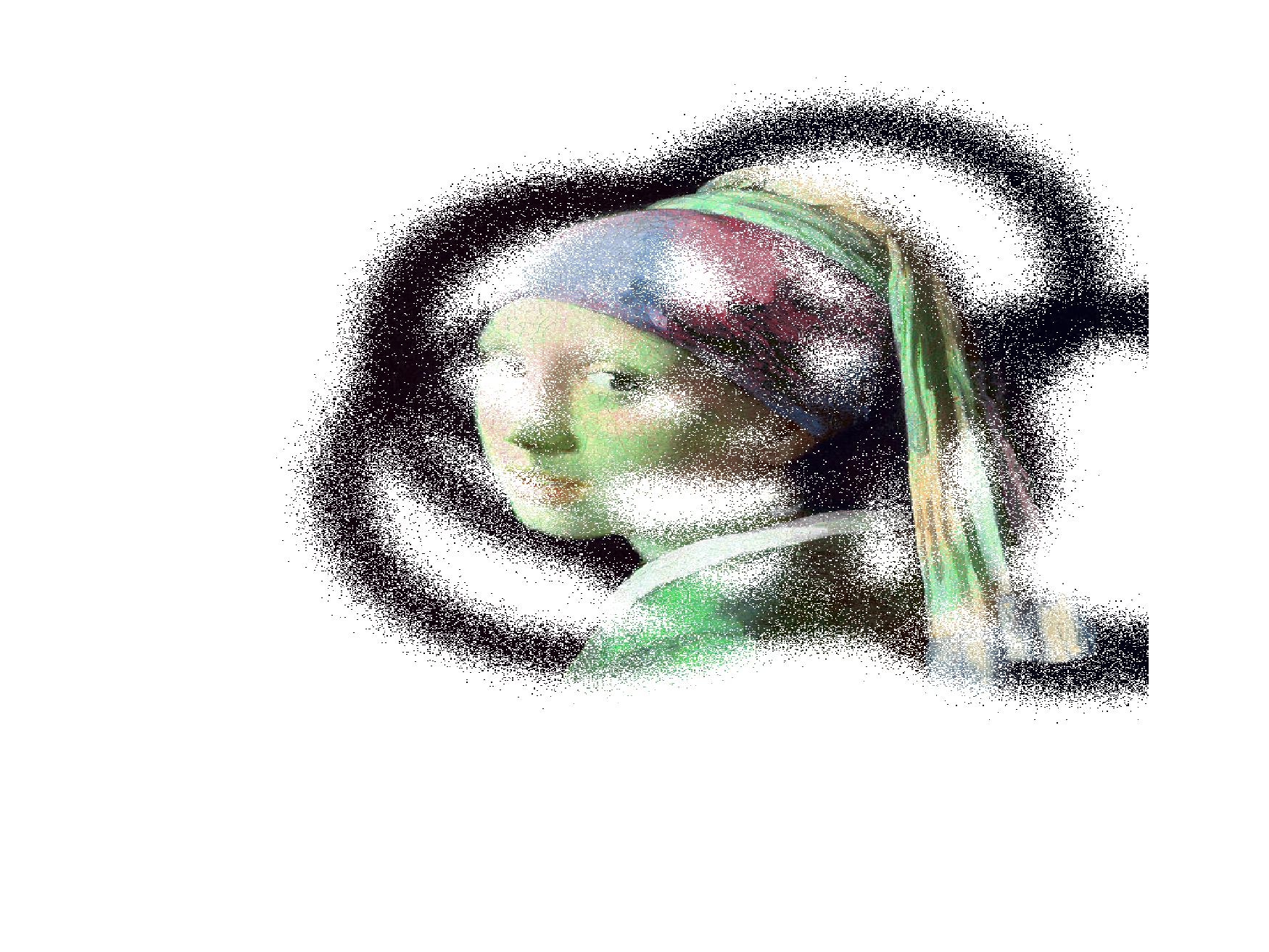}

\includegraphics[trim = 190mm 40mm 70mm 50mm,clip,width=5.75cm, height=5.75cm]{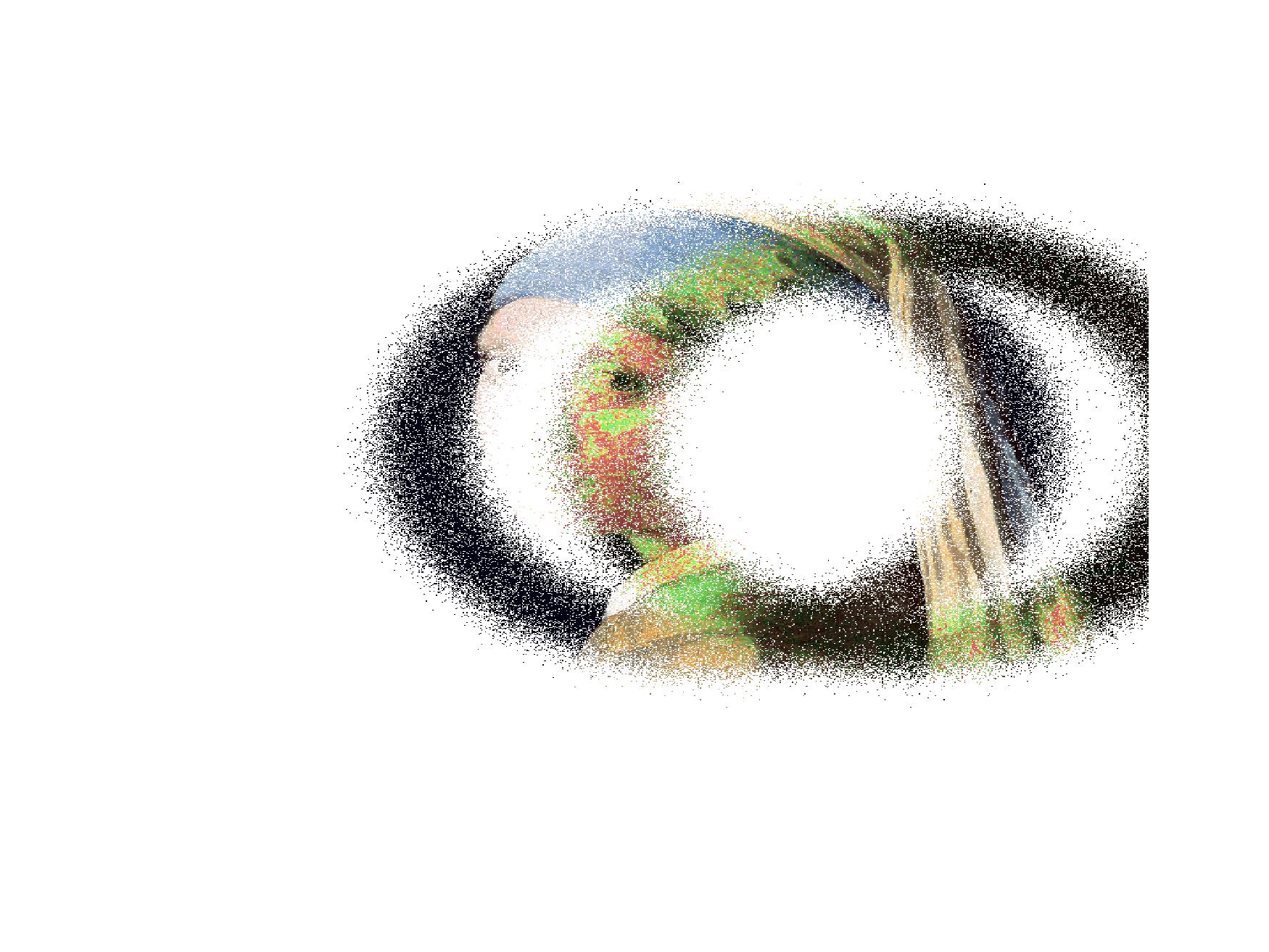}
\includegraphics[trim = 180mm 100mm 80mm 80mm,clip,width=5.75cm, height=5.75cm]{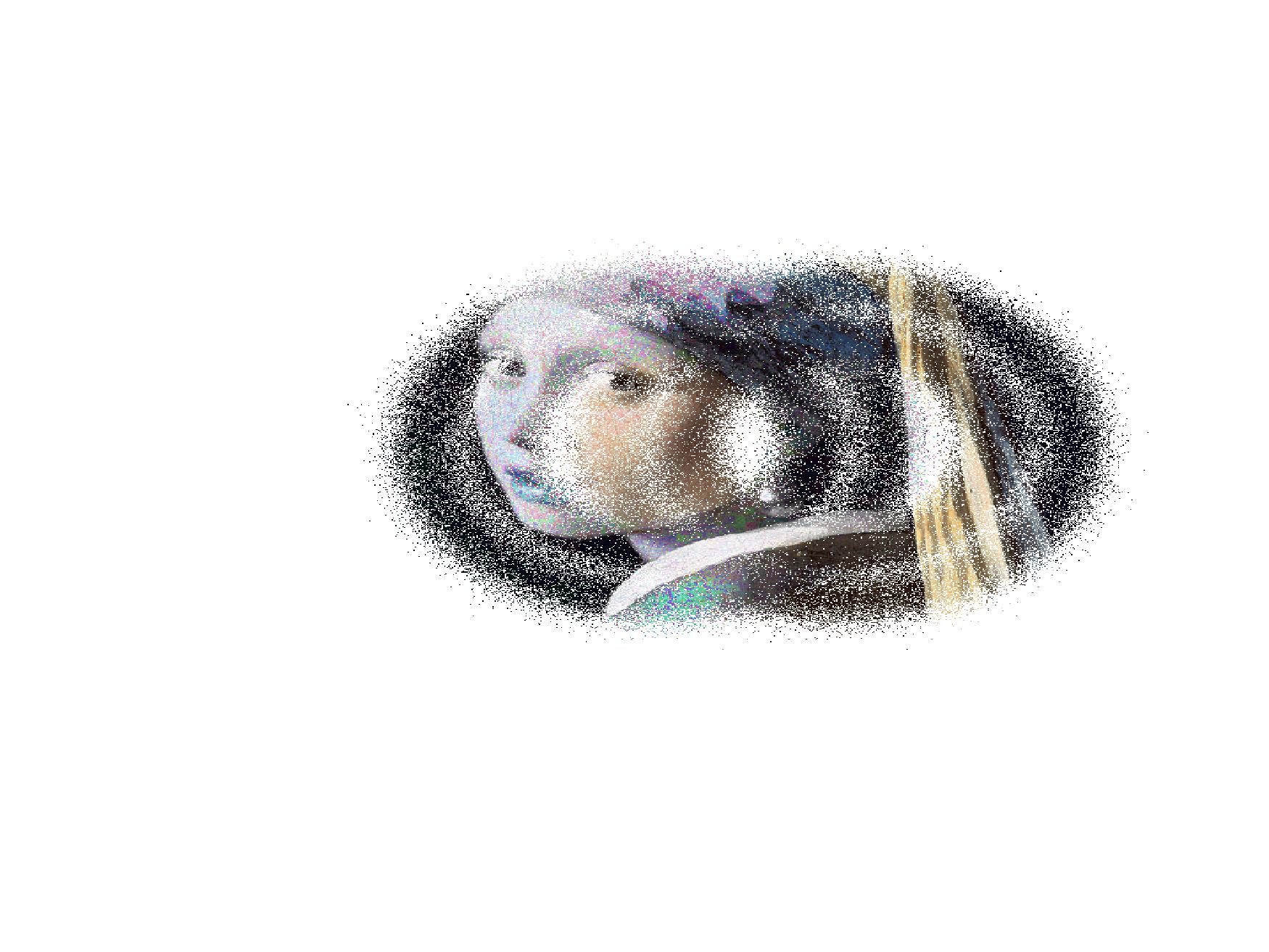}

\includegraphics[trim = 190mm 40mm 70mm 50mm,clip,width=5.75cm, height=5.75cm]{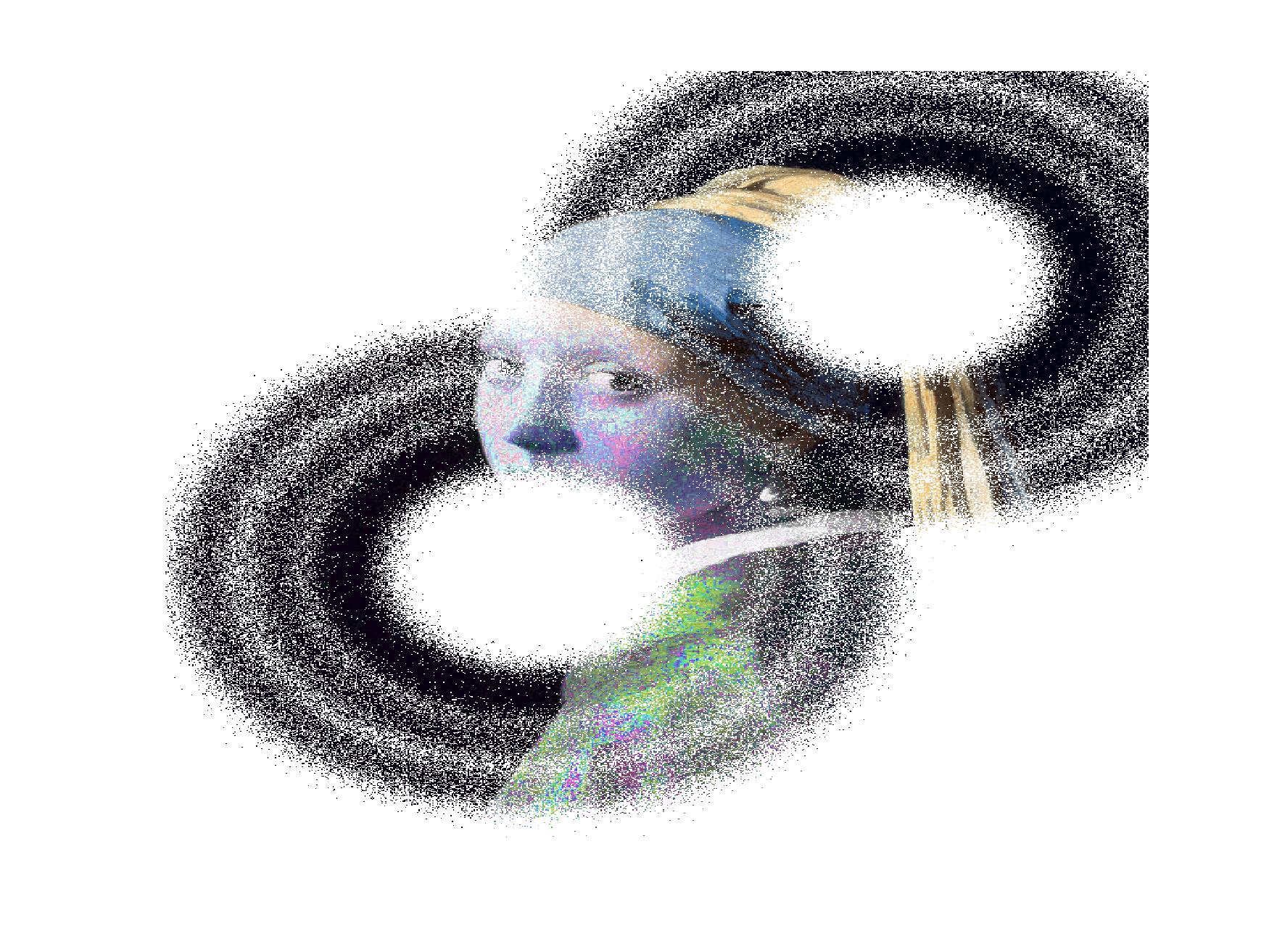}
\includegraphics[trim = 180mm 100mm 80mm 80mm,clip,width=5.75cm, height=5.75cm]{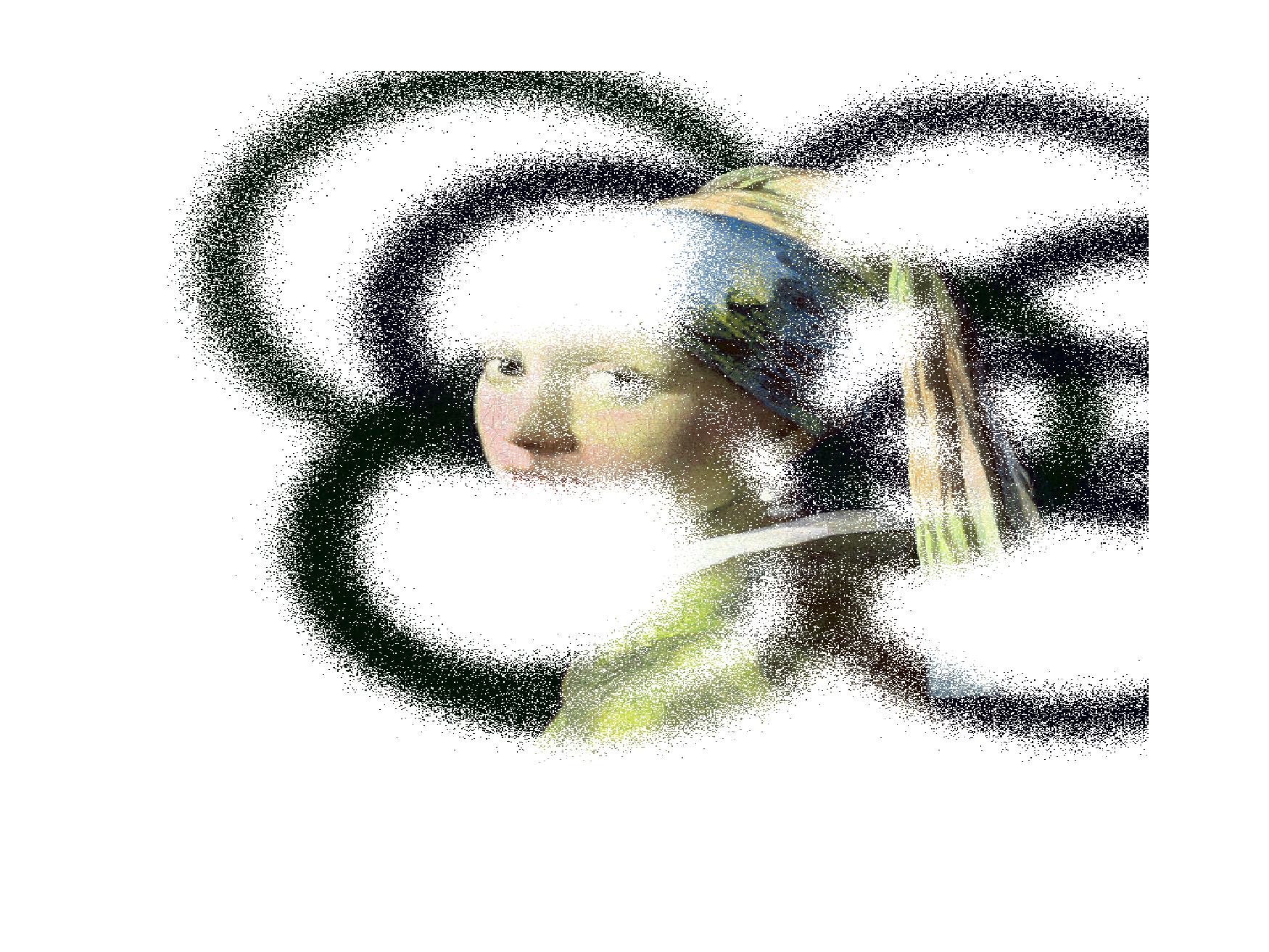}

\caption{Color symmetric art works, which can be seen  as a quotation and homage to Vermeer's {\it Girl with a Pearl Earring}. }
\label{fig:girl_stig}
\end{figure}

\begin{figure}[htb]
 
\centering
\includegraphics[trim = 80mm 40mm 70mm 0mm,clip,width=5.75cm, height=5.75cm]{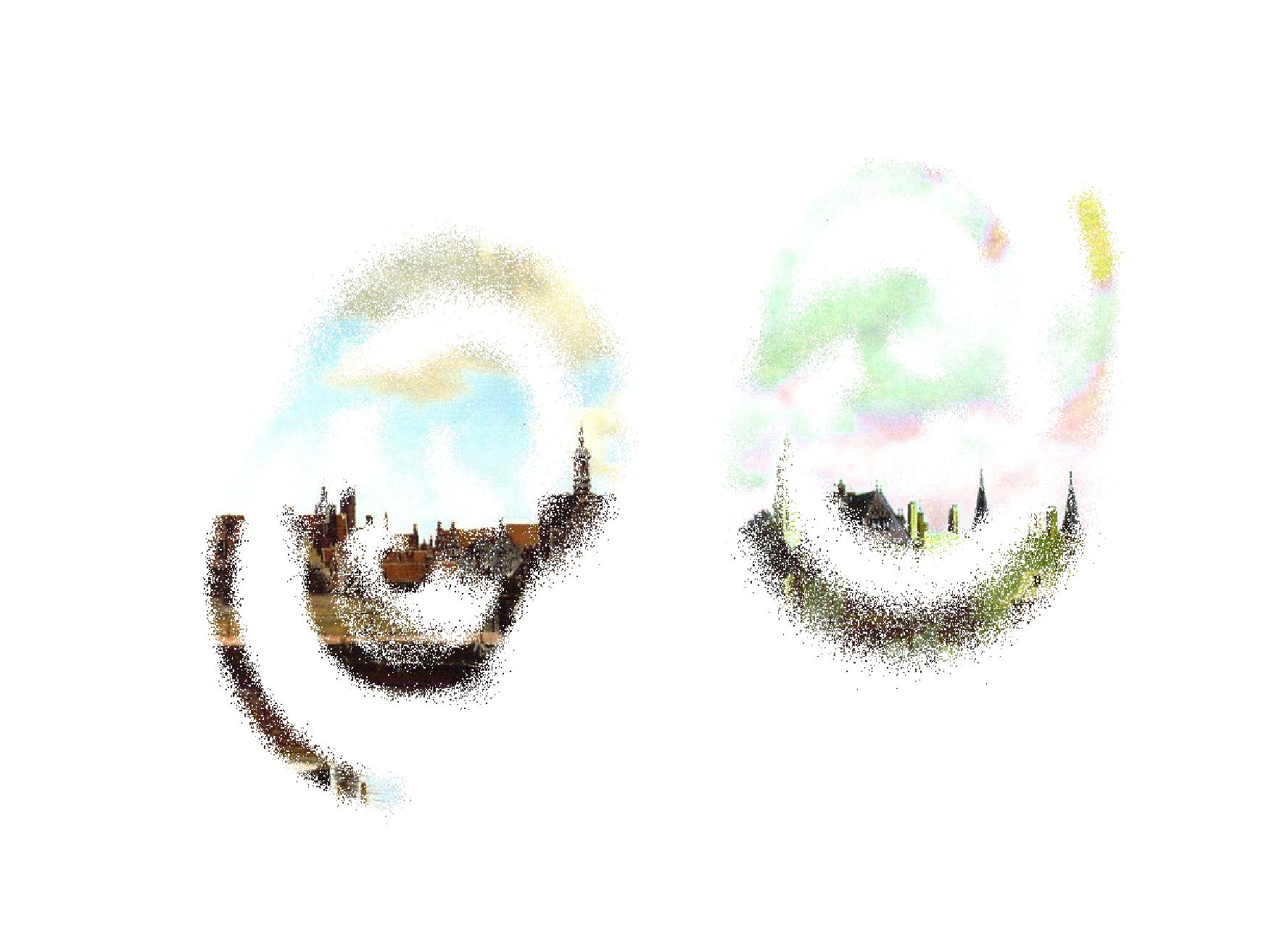}
\includegraphics[trim = 80mm 40mm 70mm 0mm,clip,width=5.75cm, height=5.75cm]{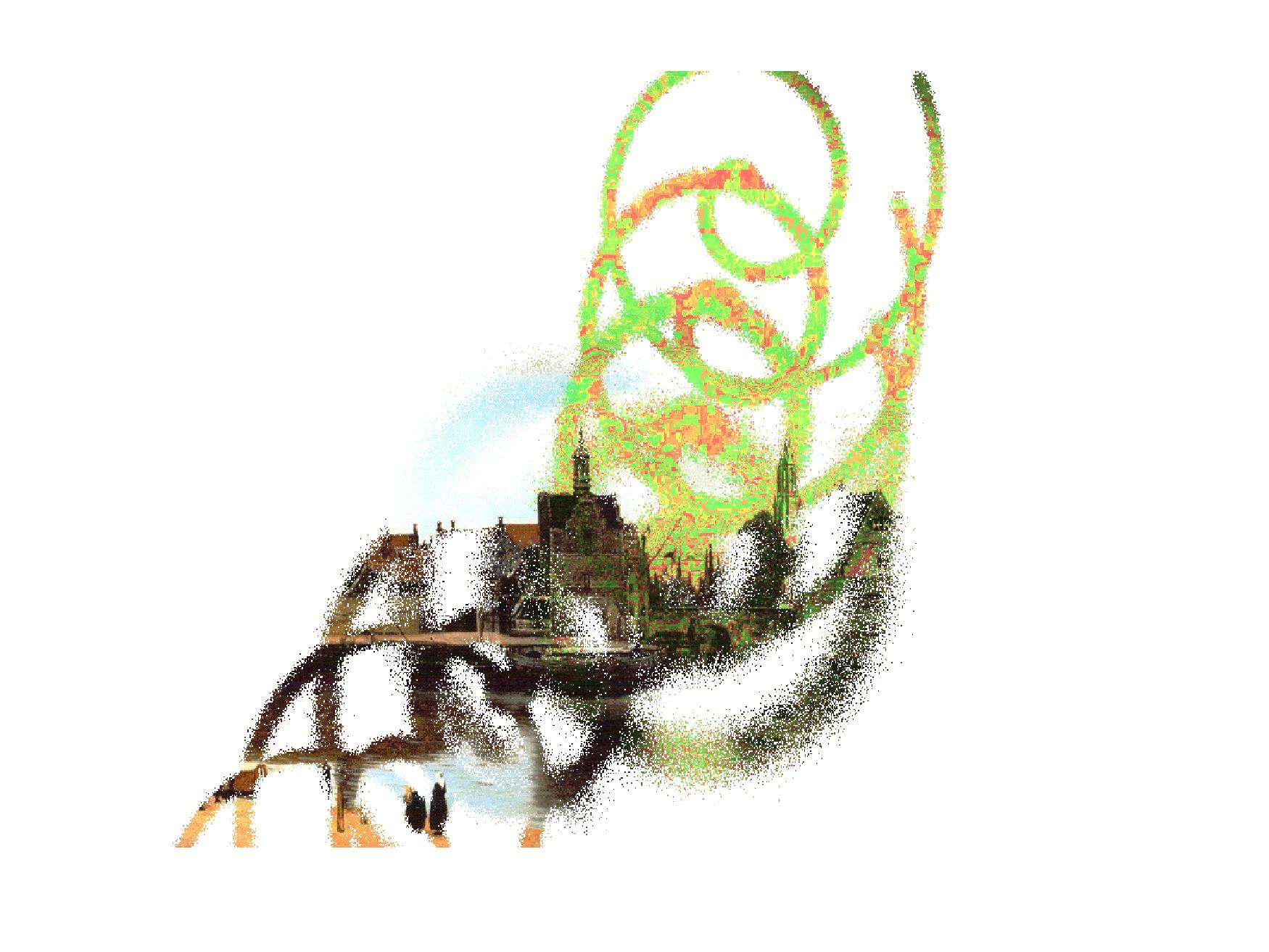}

\includegraphics[trim = 80mm 40mm 70mm 0mm,clip,width=5.75cm, height=5.75cm]{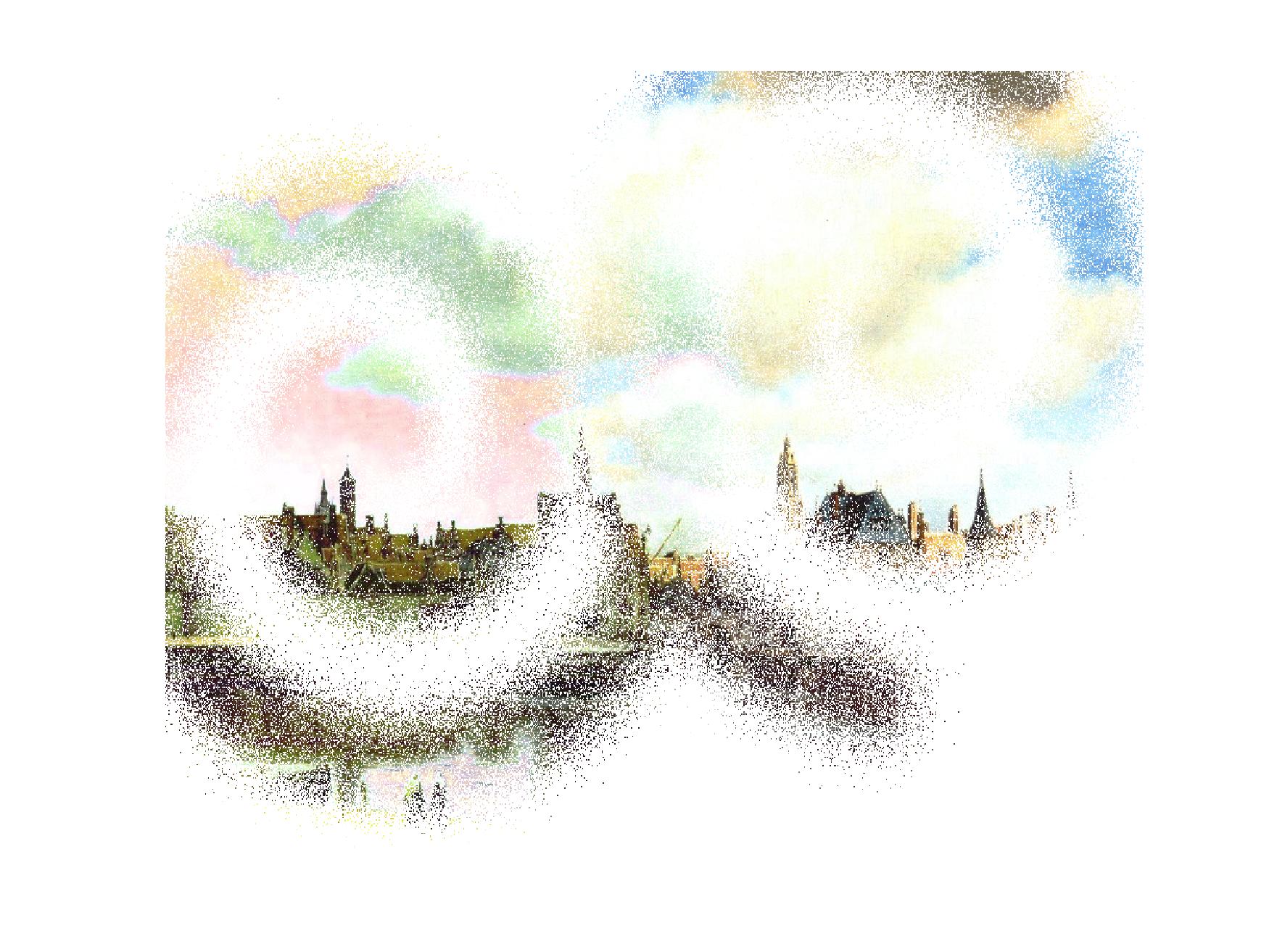}
\includegraphics[trim = 80mm 40mm 70mm 0mm,clip,width=5.75cm, height=5.75cm]{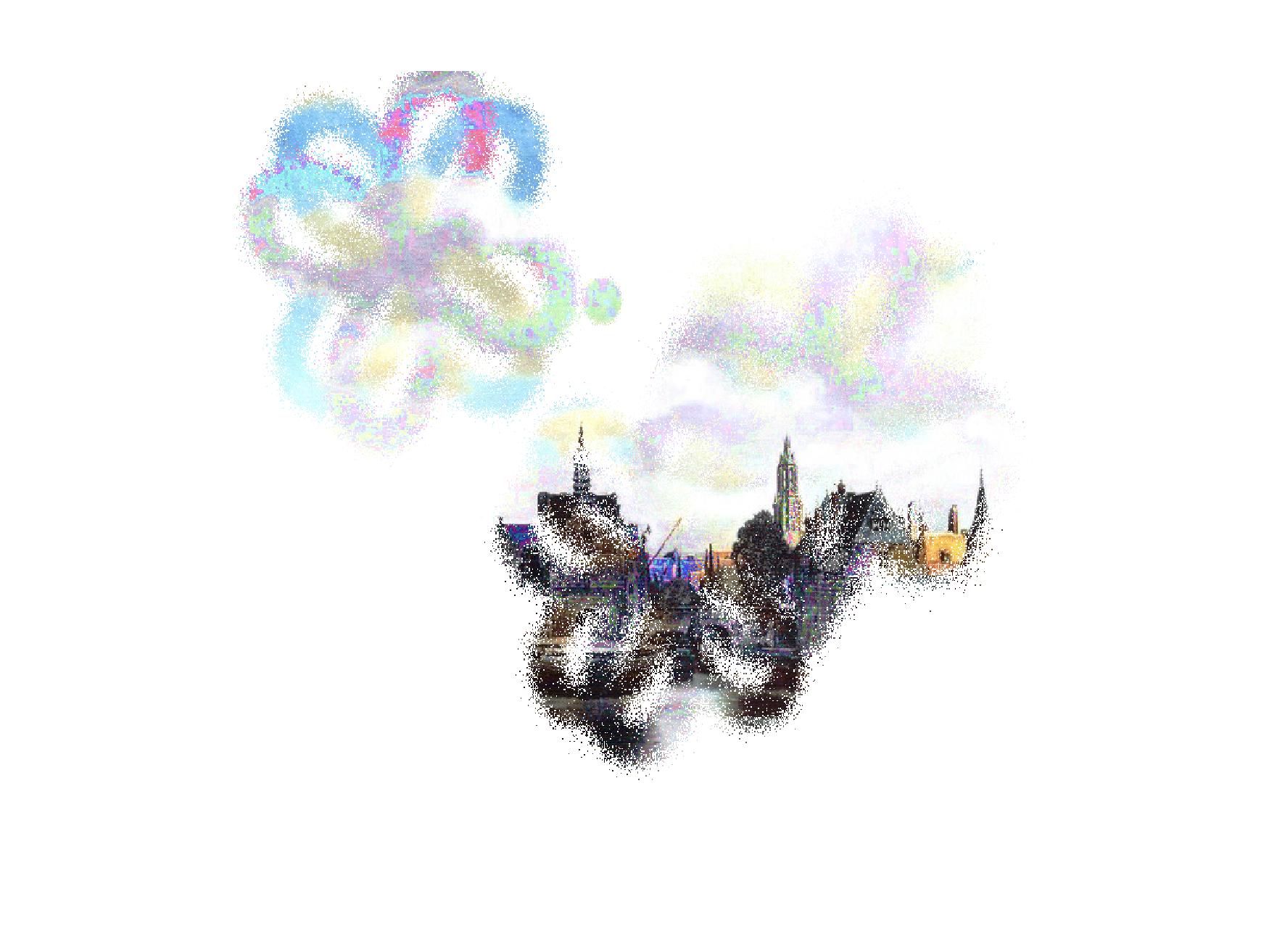}

\includegraphics[trim = 80mm 40mm 70mm 0mm,clip,width=5.75cm, height=5.75cm]{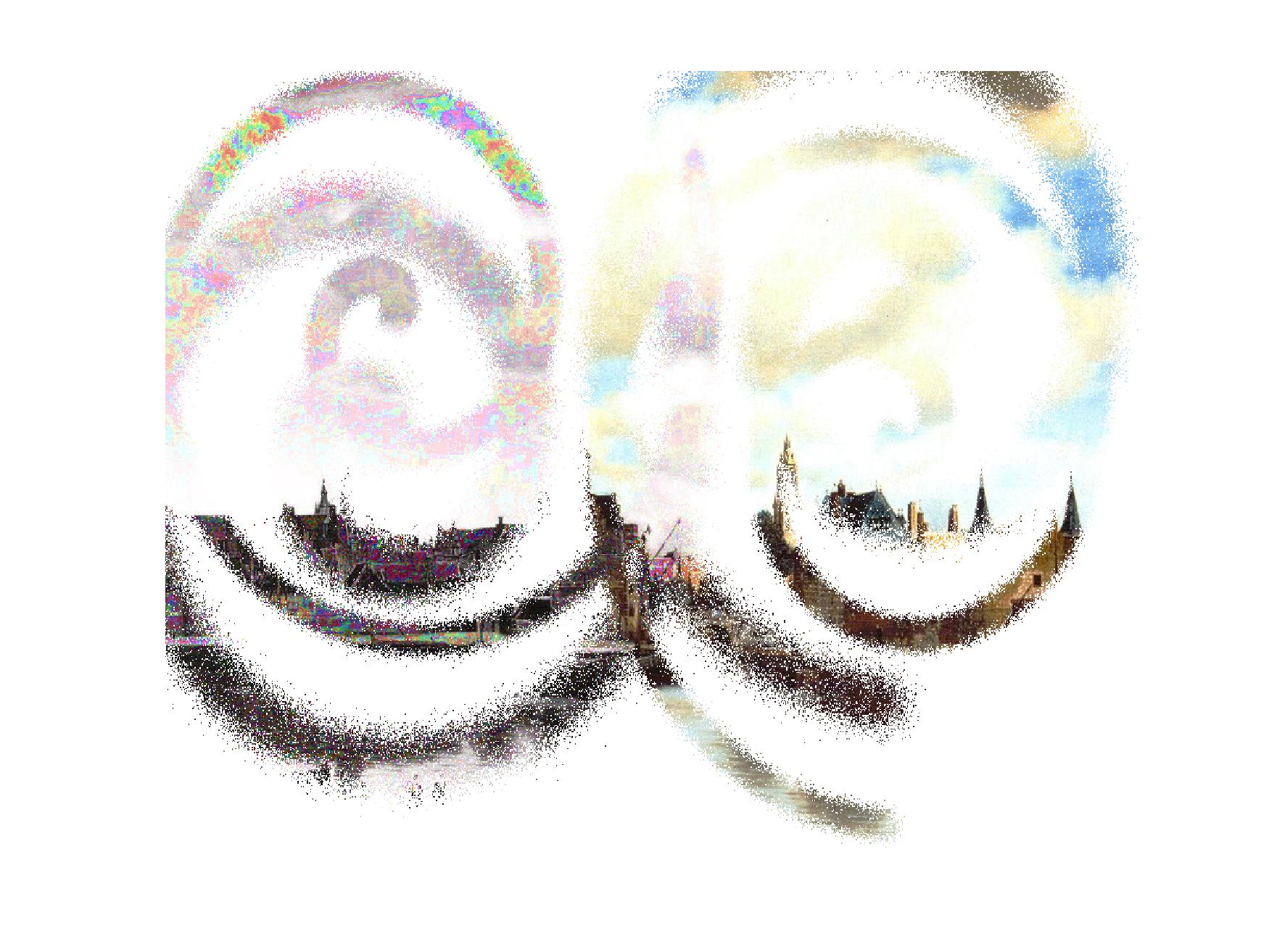}
\includegraphics[trim = 130mm 120mm 100mm 50mm,clip,width=5.75cm, height=5.75cm]{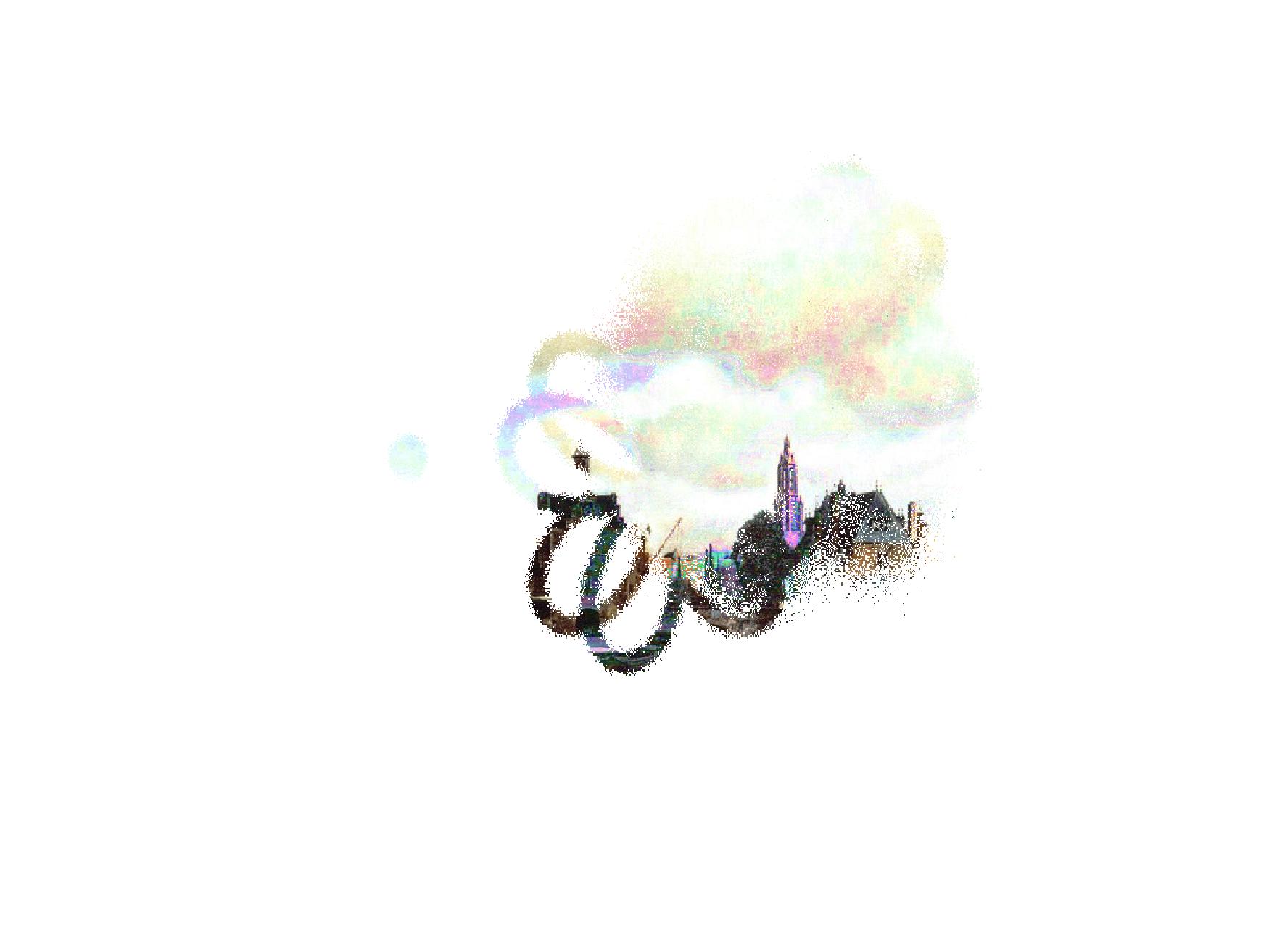}
\caption{Color symmetric art works quoting Vermeer's {\it View of Delft}.
}
\label{fig:delft_stig}
\end{figure}

A second set of experiments uses the framework of stigmergic art and color symmetry to generate images based on source images.  
Modifying  source images with the aim of creating visual art that resembles known pictures or paintings but also give a reinterpretation has also been a topic in bio-inspired generative art. For example, see the modification of 
source images from the history of art by stochastic hillclimbers, plant propagation algorithms, simulated annealing, or particle swarms~\cite{deandr20,paau19}, but also  
image transitions by random walks~\cite{alex17,neu17,neu20}.
In the experiments reported here, two well-known painting by Johannes Vermeer have been used: {\it Girl with a Pearl Earring}
(1665) and 
{\it View of Delft} (1661). (Figs. \ref{fig:vermeer}a and \ref{fig:vermeer}b). In a first step, the colors in these images are modified using color permutations similar to Eqs. \eqref{eq:col_per11} and \eqref{eq:col_per21}. In addition to the functions given above the color maps $f_4(h_i)=4h_i(h_i-1)+1$, $f_5=0.15(1+\cos{(40\pi h_i)})+0.5h_i$ and $f_k(h_i)=kh_i \: \text{mod} \: 1$ have been used, see the examples of a triangular pattern and a chess board pattern with 16 elements in Figs. \ref{fig:vermeer}c and \ref{fig:vermeer}d. 
Subsequently, patterns of nests are generated where the grains  take the colors of the source image while the grains of the symmetric nest patterns take the color-symmetric hues. Fig. \ref{fig:girl_stig} and \ref{fig:delft_stig} show results, which can be seen  as a quotation and homage to Vermeer's famous paintings. Again, the original art works remain recognizable, while at the same time an alienation or distancing effect occurs.

\clearpage

\section{Conclusion}
An important property of symmetry  is the repetitions of motifs with some features of the motif changing while some other features remain invariant. Take, for instance, a mirror image of a motif, which appears as an almost identical copy but is reversed in the direction perpendicular to the mirror surface. 
A color symmetry permutes the coloring of objects consistently with their symmetry group. Thus, it implies a symmetry-consistent coloring of  the motif and its copies. Suppose a motif adheres to a certain coloring scheme, then the mirror  image
should change or preserve the colors consistently. In this paper experiments  with and visual results of designing such color symmetries  have been discussed in bio-inspired generative art using principles from stigmergy.  
Therefore, a model of stigmergic nest construction proposed by Urbano~\cite{urb11} (see also Greenfield~\cite{green12,green14} for further usage in generative art) has been used to produce geometrical structures. These structures have been interpreted as motifs. In the visual results, the stigmergic nest motifs are colored according to algorithmic ideas exploiting symmetry and color symmetry. 

The results given in the paper only use a small subset of the design space spanned by the algorithmic framework. Thus, future work could explore the possibilities of the design space to a larger extent. For instance, the focus of color changing maps was on hue $h_i$. Therefore, different hue mappings $f(h_i)$ have been discussed while the value (brightness) $v_i$ and  the saturation $s_i$ remained constant. As the same method can be used for the whole HSV color space, it would be interesting to simultaneously change these elements of the color space for addressing different aspects of symmetry at the same time. In addition, it would be interesting to 
evaluate the results by computational aesthetics measures in order to learn more about how aesthetic feature values are modified by color-symmetric changes.

\end{document}